\documentclass[10pt,twocolumn,letterpaper]{article}

\usepackage{wacv}
\usepackage{times}
\usepackage{epsfig}
\usepackage{graphicx}
\usepackage{amsmath}
\usepackage{amssymb}
\usepackage{booktabs}
\usepackage{subcaption}
\usepackage{multirow}
\usepackage[accsupp]{axessibility}  % Improves PDF readability for those with disabilities.

\makeatletter
\@namedef{ver@everyshi.sty}{}
\makeatother
\usepackage{tikz}
\usepackage{ifthen}
\usetikzlibrary{matrix,calc}
\usepackage{xstring} % added to extract the . <<<<<

\usepackage{enumitem}
\setlist[itemize]{noitemsep, topsep=0pt}

\usepackage{pifont}
\usepackage{amsmath}
\usepackage{amssymb}
\usepackage{color,soul}

\newcommand{\mytilde}{\raise.17ex\hbox{$\scriptstyle\mathtt{\sim}$}}

\def\ll{\mathbf{l}}

\def\sss{\mathbf{s}}

\def\vv{\mathbf{v}}
\def\ww{\mathbf{w}}
\def\xx{\mathbf{x}}
\def\yy{\mathbf{y}}

\def\lL{\mathcal{L}}
\def\mM{\mathcal{M}}

\def\sS{\mathcal{S}}

\def\vV{\mathcal{V}}
\def\wW{\mathcal{W}}

\def\wacvPaperID{425} % Enter the WACV Paper ID here

\wacvapplicationstrack % Uncomment this line if you are submitting to the Applications Track.

\wacvfinalcopy % *** Uncomment this line for the final submission

\ifwacvfinal
\usepackage[breaklinks=true,bookmarks=false]{hyperref}
\else
\usepackage[pagebackref=true,breaklinks=true,colorlinks,bookmarks=false]{hyperref}
\fi

\begin{document}

%%%%%%%%% TITLE
\title{TinyHD: Efficient Video Saliency Prediction with Heterogeneous Decoders using Hierarchical Maps Distillation}

%\author{Feiyan Hu\\
%Institution1\\
%Institution1 address\\
%{\tt\small firstauthor@i1.org}
% For a paper whose authors are all at the same institution,
% omit the following lines up until the closing ``}''.
% Additional authors and addresses can be added with ``\and'',
% just like the second author.
% To save space, use either the email address or home page, not both
%\and
%Second Author\\
%Institution2\\
%First line of institution2 address\\
%{\tt\small secondauthor@i2.org}
%}
%\author{Feiyan Hu \and Simone Palazzo \and Federica Proietto Salanitri \and Giovanni Bellitto \and Morteza Moradi \and Concetto Spampinato \and Kevin McGuinness\\
%Insight SFI Research Centre for Data Analytics, Dublin City University, Dublin, Ireland\\
%PeRCeiVe Lab, University of Catania, Catania, Italty\\
%}
\author{%
  Feiyan Hu\textsuperscript{1}, Simone Palazzo\textsuperscript{2}, Federica Proietto Salanitri\textsuperscript{2}, Giovanni Bellitto\textsuperscript{2}, Morteza Moradi\textsuperscript{2}, \\ Concetto Spampinato\textsuperscript{2}, Kevin McGuinness\textsuperscript{1}\\
  %\IEEEauthorblockN{%
  %  Author One\IEEEauthorrefmark{1}\textsuperscript{\textsection},
  %  Author 2\IEEEauthorrefmark{2}\textsuperscript{\textsection},
  %  Author 3\IEEEauthorrefmark{3} and
  %  Author 4\IEEEauthorrefmark{1}%
  %}%
  \textsuperscript{1} Insight SFI Research Centre for Data Analytics, Dublin City University, Dublin, Ireland\\
  {\tt\small \{feiyan.hu, kevin.mcguinness\}@dcu.ie}\\
  \textsuperscript{2} PeRCeiVe Lab, University of Catania, Catania, Italy\\
  {\tt\small \{simone.palazzo, concetto.spampinato\}@unict.it}
  %\IEEEauthorblockA{Affiliation 1}%
  %\IEEEauthorblockA{\IEEEauthorrefmark{2} Affiliation 2}%
  %\IEEEauthorblockA{\IEEEauthorrefmark{3} Affiliation 3}%
}

\maketitle
\thispagestyle{empty}

%%%%%%%%% ABSTRACT
\begin{abstract}
   Video saliency prediction has recently attracted attention of the research community, as it is an upstream task for several practical applications. However, current solutions are particularly computationally demanding, especially due to the wide usage of spatio-temporal 3D convolutions. We observe that, while different model architectures achieve similar performance on benchmarks, visual variations between predicted saliency maps are still significant. Inspired by this intuition, we propose a lightweight model that employs multiple simple heterogeneous decoders and adopts several practical approaches to improve accuracy while keeping computational costs low, such as hierarchical multi-map knowledge distillation, multi-output saliency prediction, unlabeled auxiliary datasets and channel reduction with teacher assistant supervision. Our approach achieves saliency prediction accuracy on par or better than state-of-the-art methods on DFH1K, UCF-Sports and Hollywood2 benchmarks, while enhancing significantly the efficiency of the model.
\end{abstract}

\section{Introduction}
Video saliency prediction aims at estimating patterns of human attention during free-viewing of dynamic scenes, to emulate the capabilities of the human visual system of quickly analyzing and interpreting the surrounding environment. Due to its several practical applications \cite{fang2021dada,droste2019towards,perrin2019well,sun2017camera,6784518,zhang2009visual,hadizadeh2013saliency,10.1145/3313831.3376544}, it is an active area of research in computer vision. However, the solution to this problem is not trivial, for several reasons. First, attention mechanisms in the human visual system are not fully known, so it is not clear how to emulate them. Also, it requires complex modeling of both visual features and their motion and interaction: an object with striking visual patterns may be shadowed by a bland element of the scene that starts moving in a peculiar way. Finally, modeling the temporal dimension may become computationally expensive, especially with current deep learning methods based on spatio-temporal 3D convolutions, thus limiting the applicability to low-power devices.

%\begin{figure}
%  \centering
%  \begin{subfigure}{0.49\linewidth}
%   \includegraphics[width=\textwidth, trim={0 0.5cm 1cm 1.3cm},clip]{figures/seaborn_test_cc.pdf}
%    \caption{CC metric}
%    \label{fig:cc_soa}
%  \end{subfigure}
%  \hspace{-0.9\baselineskip}
%  \begin{subfigure}{0.49\linewidth}
%   \includegraphics[width=\textwidth, trim={0 0.5cm 1cm 1.3cm},clip]{figures/seaborn_test_sim.pdf}
%    \caption{SIM metric}
%    \label{fig:sim_soa}
%  \end{subfigure}
%\caption{Measuring prediction similarity among video saliency prediction models TASED, ViNet and HD2S on DHF1K validation set.}
%\label{fig:similarity_soa}
%\end{figure}
    \def\myConfMat{% divided by 10
        { 
            {100,84,72},  %row 1
            {84,100,75},  %row 2
            {72,75,100},  %row 3
    }}

    \def\myConfMatsim{% divided by 10
        { 
            {100,73,57},  %row 1
            {73,100,58},  %row 2
            {57,58,100},  %row 3
    }}

    \def\classNames{{"TASED","ViNet","HD2S"}} %class names. Adapt at will  

    \def\numClasses{3} %number of classes. Could be automatic, but you can change it for tests.
    
    \def\myScale{0.49} % 1.5 is a good scale. Values under 1 may need smaller fonts!
    
\begin{figure}
    \centering
    \begin{tikzpicture}[
        scale = \myScale,
        %font={\scriptsize}, %for smaller scales, even \tiny may be useful
        font={\tiny}, %for smaller scales, even \tiny may be useful
        ]
        
        \tikzset{vertical label/.style={rotate=90,anchor=east}}   % usable styles for below
        \tikzset{diagonal label/.style={rotate=45,anchor=north east}}
        
        \foreach \y in {1,...,\numClasses} %loop vertical starting on top
        {
            % Add class name on the left
            \node [anchor=east] at (0.4,-\y) {\pgfmathparse{\classNames[\y-1]}\pgfmathresult};          
            \foreach \x in {1,...,\numClasses}  %loop horizontal starting on left
            {
                %---- Start of automatic calculation of totSamples for the column ------------  
                \pgfmathsetmacro{\totSamples}{0}
                \foreach \ll in {1,...,\numClasses}
                {
                      \pgfmathsetmacro{\tmp}{\totSamples+ \myConfMat[\ll-1][\x-1]} %accumulate it with previous
                      \global\let\totSamples\tmp% put the final sum in variable
                }
                %---- End of automatic calculation of totSamples ----------------               
                \begin{scope}[shift={(\x,-\y)}]
                    \pgfmathsetmacro{\r}{\myConfMat[\y-1][\x-1]}   %                            
                    \pgfmathtruncatemacro{\p}{round(\r-70)}
                    %\pgfmathtruncatemacro{\p}{\myConfMat[\y-1][\x-1]}
                    \coordinate (C) at (0,0);
                    \ifthenelse{\p<50}{\def\txtcol{black}}{\def\txtcol{white}} %decide text color for contrast
                    \node[
                    draw,                 %draw lines
                    text=\txtcol,         %text color (automatic for better contrast)
                    align=center,         %align text inside cells (also for wrapping)
                    fill=blue!\p,        %intensity of fill (can change base color)
                    minimum size=\myScale*10mm,    %cell size to fit the scale and integer dimensions (in cm)
                    inner sep=0,          %remove all inner gaps to save space in small scales
                    ] (C) {\r\%}; %{\StrSubstitute{\r}{.}{}\\\p\%};     %text to put in cell (adapt at will)
                    %Now if last vertical class add its label at the bottom
                    \ifthenelse{\y=\numClasses}{
                        \node [] at ($(C)-(0,0.75)$) % can use vertical or diagonal label as option
                        {\pgfmathparse{\classNames[\x-1]}\pgfmathresult};}{}
                \end{scope}
            }
        }
        %Now add x and y labels on suitable coordinates
        \coordinate (yaxis) at (-0.85,0.5-\numClasses/2);  %must adapt if class labels are wider!
        \coordinate (xaxis) at (0.5+\numClasses/2, -\numClasses-1.25); %id. for non horizontal labels!
        %\node [vertical label] at (yaxis) {Actual Class};
        \node []               at (xaxis) {CC Metric};
    \end{tikzpicture}
    \hspace{2.5\baselineskip}
    \begin{tikzpicture}[
        scale = \myScale,
        %font={\scriptsize}, %for smaller scales, even \tiny may be useful
        font={\tiny}, %for smaller scales, even \tiny may be useful
        ]
        
        \tikzset{vertical label/.style={rotate=90,anchor=east}}   % usable styles for below
        \tikzset{diagonal label/.style={rotate=45,anchor=north east}}
        
        \foreach \y in {1,...,\numClasses} %loop vertical starting on top
        {
            % Add class name on the left
            \node [anchor=east] at (0.4,-\y) {\pgfmathparse{\classNames[\y-1]}\pgfmathresult};          
            \foreach \x in {1,...,\numClasses}  %loop horizontal starting on left
            {
                %---- Start of automatic calculation of totSamples for the column ------------  
                \pgfmathsetmacro{\totSamples}{0}
                \foreach \ll in {1,...,\numClasses}
                {
                      \pgfmathsetmacro{\tmp}{\totSamples+ \myConfMatsim[\ll-1][\x-1]} %accumulate it with previous
                      \global\let\totSamples\tmp% put the final sum in variable
                }
                %---- End of automatic calculation of totSamples ----------------               
                \begin{scope}[shift={(\x,-\y)}]
                    \pgfmathsetmacro{\r}{\myConfMatsim[\y-1][\x-1]}   %                            
                    \pgfmathtruncatemacro{\p}{round(\r-70)}
                    %\pgfmathtruncatemacro{\p}{\myConfMat[\y-1][\x-1]}
                    \coordinate (C) at (0,0);
                    \ifthenelse{\p<50}{\def\txtcol{black}}{\def\txtcol{white}} %decide text color for contrast
                    \node[
                    draw,                 %draw lines
                    text=\txtcol,         %text color (automatic for better contrast)
                    align=center,         %align text inside cells (also for wrapping)
                    fill=blue!\p,        %intensity of fill (can change base color)
                    minimum size=\myScale*10mm,    %cell size to fit the scale and integer dimensions (in cm)
                    inner sep=0,          %remove all inner gaps to save space in small scales
                    ] (C) {\r\%}; %{\StrSubstitute{\r}{.}{}\\\p\%};     %text to put in cell (adapt at will)
                    %Now if last vertical class add its label at the bottom
                    \ifthenelse{\y=\numClasses}{
                        \node [] at ($(C)-(0,0.75)$) % can use vertical or diagonal label as option
                        {\pgfmathparse{\classNames[\x-1]}\pgfmathresult};}{}
                \end{scope}
            }
        }
        %Now add x and y labels on suitable coordinates
        \coordinate (yaxis) at (-0.85,0.5-\numClasses/2);  %must adapt if class labels are wider!
        \coordinate (xaxis) at (0.5+\numClasses/2, -\numClasses-1.25); %id. for non horizontal labels!
        %\node [vertical label] at (yaxis) {Actual Class};
        \node []               at (xaxis) {SIM Metric};
    \end{tikzpicture}
\caption{Measuring prediction similarity among video saliency prediction models TASED, ViNet and HD2S on DHF1K validation set.}
\label{fig:similarity_soa}
\end{figure}

\begin{figure*}
	\centering
	\begin{subfigure}[b]{0.13\textwidth}
		
		\includegraphics[width=\textwidth]{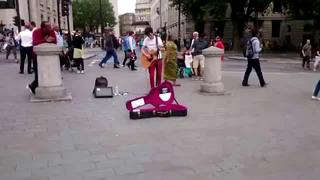}
%		\caption{$y=x$}
		\label{a_1}
	\end{subfigure}
	\begin{subfigure}[b]{0.13\textwidth}
		
		\includegraphics[width=\textwidth]{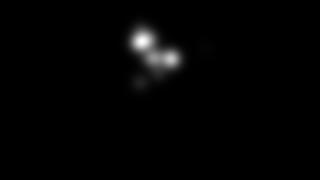}
%		\caption{$y=3sinx$}
		\label{b_1}
	\end{subfigure}
\begin{subfigure}[b]{0.13\textwidth}
	
	\includegraphics[width=\textwidth]{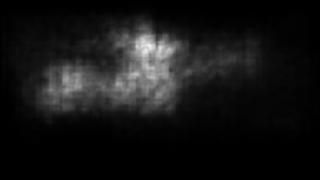}
%	\caption{$y=3sinx$}
	\label{c_1}
\end{subfigure}
\begin{subfigure}[b]{0.13\textwidth}
	
	\includegraphics[width=\textwidth]{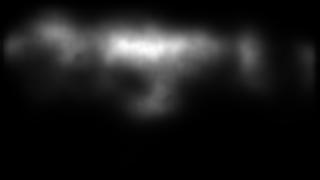}
%	\caption{$y=3sinx$}
	\label{d_1}
\end{subfigure}
\begin{subfigure}[b]{0.13\textwidth}

	\includegraphics[width=\textwidth]{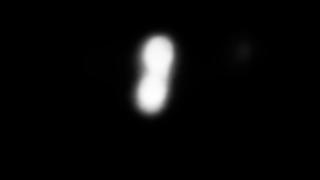}
%	\caption{$y=3sinx$}
	\label{e_1}
\end{subfigure}
	\begin{subfigure}[b]{0.13\textwidth}

		\includegraphics[width=\textwidth]{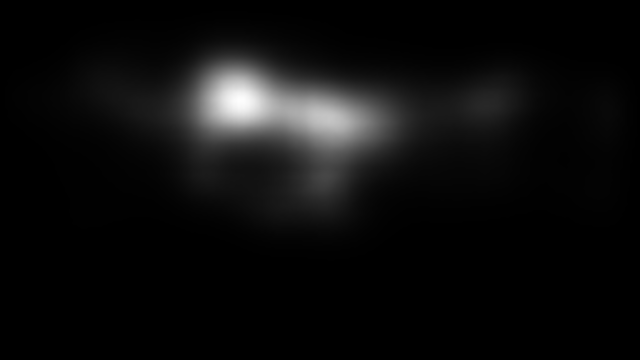}
%		\caption{$y=5/x$}
		\label{f_1}
	\end{subfigure}
	
\vspace{-0.9\baselineskip}
\begin{subfigure}[b]{0.13\textwidth}
	
	\includegraphics[width=\textwidth]{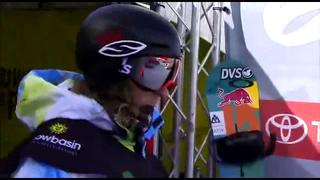}
%	\caption{$y=x$}
	\label{a_2}
\end{subfigure}
\begin{subfigure}[b]{0.13\textwidth}
	
	\includegraphics[width=\textwidth]{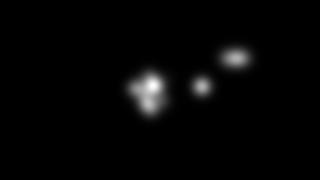}
%	\caption{$y=3sinx$}
	\label{b_2}
\end{subfigure}
\begin{subfigure}[b]{0.13\textwidth}
	
	\includegraphics[width=\textwidth]{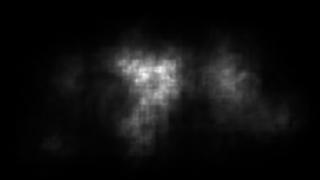}
%	\caption{$y=3sinx$}
	\label{c_2}
\end{subfigure}
\begin{subfigure}[b]{0.13\textwidth}
	
	\includegraphics[width=\textwidth]{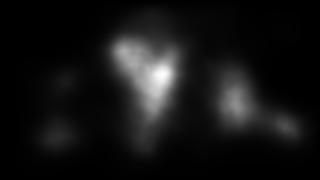}
%	\caption{$y=3sinx$}
	\label{d_2}
\end{subfigure}
\begin{subfigure}[b]{0.13\textwidth}
	
	\includegraphics[width=\textwidth]{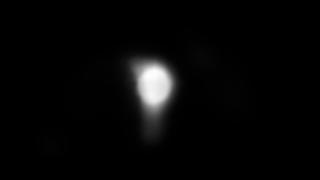}
%	\caption{$y=3sinx$}
	\label{e_2}
\end{subfigure}
\begin{subfigure}[b]{0.13\textwidth}
	\includegraphics[width=\textwidth]{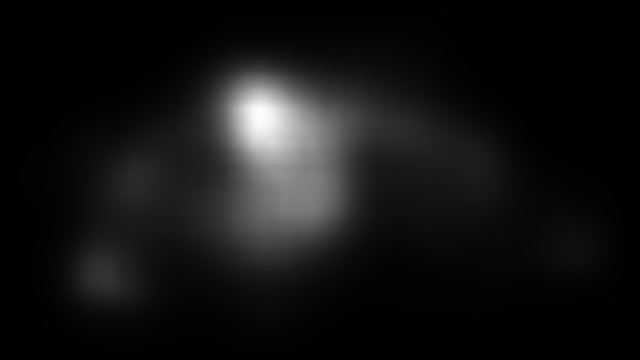}
%	\caption{$y=5/x$}
	\label{f_2}
\end{subfigure}

\vspace{-0.9\baselineskip}
\begin{subfigure}[b]{0.13\textwidth}
	
	\includegraphics[width=\textwidth]{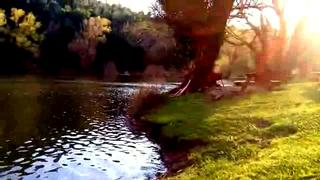}
	%	\caption{$y=x$}
	\caption{Frame}
	\label{a_3}
\end{subfigure}
\begin{subfigure}[b]{0.13\textwidth}
	
	\includegraphics[width=\textwidth]{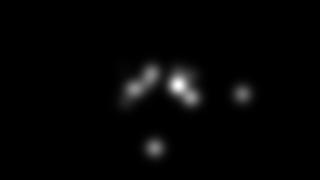}
	%	\caption{$y=3sinx$}
	\caption{GT}
	\label{b_3}
\end{subfigure}
\begin{subfigure}[b]{0.13\textwidth}
	
	\includegraphics[width=\textwidth]{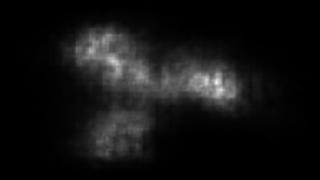}
	%	\caption{$y=3sinx$}
	\caption{TASED}
	\label{c_3}
\end{subfigure}
\begin{subfigure}[b]{0.13\textwidth}
	
	\includegraphics[width=\textwidth]{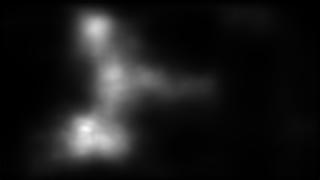}
	%	\caption{$y=3sinx$}
	\caption{ViNet}
	\label{d_3}
\end{subfigure}
\begin{subfigure}[b]{0.13\textwidth}
	
	\includegraphics[width=\textwidth]{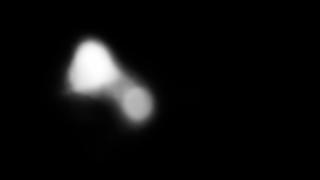}
	%	\caption{$y=3sinx$}
	\caption{HD2S}
	\label{e_3}
\end{subfigure}
\begin{subfigure}[b]{0.13\textwidth}

	\includegraphics[width=\textwidth]{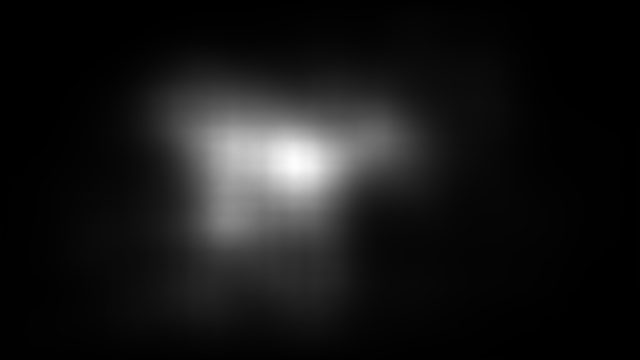}
	%	\caption{$y=5/x$}
	\caption{DLA}
	\label{f_3}
\end{subfigure}

%\vspace{-0.9\baselineskip}
%\begin{subfigure}[b]{0.16\textwidth}
	
%	\includegraphics[width=\textwidth]{figures/vid_0070_0696.jpg}
%	\caption{Frame}
%	\label{a_4}
%\end{subfigure}
%\begin{subfigure}[b]{0.16\textwidth}
	
%	\includegraphics[width=\textwidth]{figures/gt_0070_0696.jpg}
%	\caption{GT}
%	\label{b_4}
%\end{subfigure}
%\begin{subfigure}[b]{0.16\textwidth}
	
%	\includegraphics[width=\textwidth]{figures/tased_0070_0696.jpg}
%	\caption{TASED}
%	\label{c_4}
%\end{subfigure}
%\begin{subfigure}[b]{0.16\textwidth}
	
%	\includegraphics[width=\textwidth]{figures/vinet_0070_0696.jpg}
%	\caption{ViNet}
%	\label{d_4}
%\end{subfigure}
%\begin{subfigure}[b]{0.16\textwidth}
	
%	\includegraphics[width=\textwidth]{figures/hd2s_0070_0696.jpg}
%	\caption{HD2S}
%	\label{e_4}
%\end{subfigure}
%\begin{subfigure}[b]{0.16\textwidth}
	%\includegraphics[width=\textwidth]{figures/dla_0070_0696.jpg}
%	\includegraphics[width=\textwidth]{figures/s3ddla_0070_0696.png}
%	\caption{DLA}
%	\label{f_4}
%\end{subfigure}
	\caption{Examples of video saliency maps from state-of-the-art methods. Although they achieve very similar performance on popular metrics, remarkable differences can be seem in the learned saliency patterns.}
	\label{fig:saliency_example_attend_roi}
\end{figure*}

Many solutions have been proposed, based on different assumptions on how to capture video saliency. It is interesting to note that, in spite of the remarkably different research directions followed by the variety of works in the literature, top results over video saliency prediction benchmarks are very close~\cite{bellitto2021hierarchical,chang2021temporal,wang2019revisiting}, suggesting that predictions of different models are similar. We assessed the validity of this conclusion
%However, this assumption seems to be far from the truth. To demonstrate this claim, we compared
by comparing three of the best performing methods on the DHF1K dataset~\cite{wang2019revisiting} --- TASED~\cite{min2019tased}, HD2S~\cite{bellitto2021hierarchical} and ViNet \cite{jain2020vinet} --- not in terms of their scores on summary metrics, but in terms of the relative similarity of the predicted saliency maps. To illustrate our findings, Fig.~\ref{fig:similarity_soa} shows pairwise similarities between predicted maps over two common metrics, Linear Correlation Coefficient (CC) and Similarity (SIM). Although the three approaches achieve similar scores on both metrics on DHF1K (between 0.470 and 0.511 for CC, and between 0.361 and 0.406 for SIM), the same metrics computed between each other are relatively low, compared to what one would expect given their similarity to the saliency ground truth. A visual inspection of the saliency maps generated by methods under comparison confirms this behavior: Fig.~\ref{fig:saliency_example_attend_roi} shows that it is common to find cases where each approach produces remarkably different saliency maps.

Notably, all three methods --- TASED, HD2S and ViNet --- are encoder-decoder networks and share the same encoder, S3D~\cite{xie2018rethinking}, while employing different decoding strategies (a U-Net--like approach for TASED and ViNet, a hierarchical map aggregation for HD2S). This suggests that a key factor underlying differences between current video saliency prediction approaches lies in the way encoded features are processed in the decoding path, leading to models that learn specific (and often exclusive) representations. We strengthen this hypothesis by experimenting with another decoding strategy, exemplified by DLA~\cite{yu2018deep}, which combines hierarchical decoding with complex feature interactions: the results, also included in Fig.~\ref{fig:saliency_example_attend_roi}, show yet another saliency prediction pattern, while using the same encoder network, S3D. These results lead us to hypothesize that different model architectures introduce different inductive biases, which are more suitable to recognize certain patterns more than others, thus requiring to increase model capacity in order to generalize well to multiple saliency dynamics. Indeed, the size of weights of models in our analysis range between 82 MB and 116 MB, and the size of the top ten models in the DHF1K leaderboard\footnote{\url{https://mmcheng.net/videosal/}} is on average 238 MB.

Given these premises, instead of increasing the complexity of a single decoding strategy, it may be more efficient to employ multiple simpler architectures with fewer parameters, relying on each architecture's capability to attend to different salient regions and combining their results. Hence, we propose \textbf{TinyHD}, a \emph{lightweight, efficient and heterogeneous multi-decoder architecture} for video saliency prediction. The proposed method is inspired by encoder-decoder architectures, but introduces the adoption of heterogeneous decoding strategies in order to reduce the complexity of each decoder, increasing efficiency (the weights of the resulting model take only 16 MB) and improving the accuracy of predictions, as we show in our experiments.
Furthermore, along the direction of reducing computational costs while retaining high accuracy, we also introduce a novel knowledge distillation approach, based on exploiting a teacher with multiple hierarchical predictions: this allows the model to freely learn its own features, since no explicit conditioning on representations is enforced, while at the same time receiving a supervision signal that encodes information at different layers of abstraction.

Experiments confirm that our model can generate high-quality predictions with low computational costs and model size (only 16 MB). We assess the impact of our heterogeneous multi-decoder strategy by carrying out extensive ablation studies and comparing alternative architectures. We also demonstrate the effectiveness of our knowledge distillation strategy, compared to the employment of a non-hierarchical teacher.
To summarize our contributions:
\begin{itemize}
 \item We propose a decoding strategy for video saliency prediction which combines heterogeneous decoders to exploit the specific pattern analysis capabilities, while reducing the overall model complexity. To our knowledge, we are first to propose multiple saliency maps output using 3D CNN to improve model efficiency.
 \item We employ a knowledge distillation approach based on a hierarchical teacher, providing saliency maps estimated from different abstraction layers.
 \item Extensive experiments show that our model achieves state-of-the-art performance on the DHF1K benchmark, at lower computational costs of current methods. Ablation studies support the motivations for our decoding and knowledge distillation strategies.
\end{itemize}
%-------------------------------------------------------------------------
\section{Related Work}
\label{sec:related_work}
The main contributions of the proposed approach consist of a novel heterogeneous multi-decoder scheme, which combines lightweight versions of common decoding strategies, and a multi-objective knowledge distillation approach. In this section, we briefly present the state-of-the-art on these topics.

\noindent\textbf{Decoding strategies for video saliency prediction.} 
Among recent methods from the state-of-the-art for video saliency prediction, leveraging encoder-decoder networks can be considered a mainstream approach; however, %while many of the proposed methods share a similar architecture for feature encoding (for example, by using a pre-trained action recognition backbone~\cite{min2019tased,zou2021sta3d,bellitto2021hierarchical}),
several architectural variations have been proposed for feature sharing between encoder and decoder and for output reconstruction.
\begin{figure}
  \centering
  \begin{subfigure}{0.48\linewidth}
  %trim={<left> <lower> <right> <upper>}
   \includegraphics[width=\textwidth, trim={0 1cm 0 0},clip]{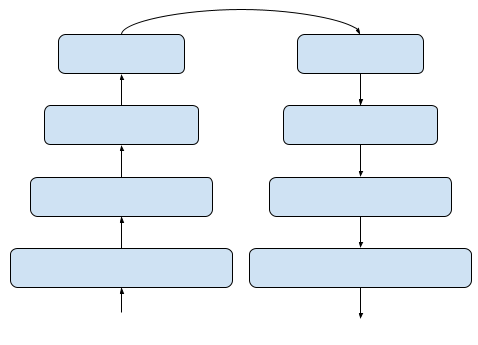}
    %\caption{Indepedent encoder and decoder, with no feature sharing between the two paths~\cite{djilali20203dsal,lai2019video,linardos2019simple,wu2020salsac,zou2021sta3d}.}
    \label{fig:naive_decoder}
  \end{subfigure}
  \hfill
  \begin{subfigure}{0.48\linewidth}
   \includegraphics[width=\textwidth, trim={0 1cm 0 0},clip]{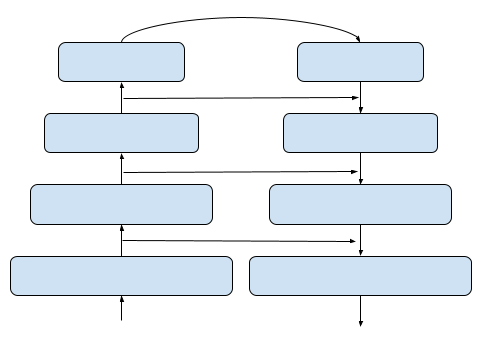}
    %\caption{Shallow decoder or U-net like decoder.}
    %\caption{U-Net--like architecture, with features sharing between encoder and decoder~\cite{droste2020unified,jain2020vinet,li2021novel,min2019tased}.} %Refer to as \textbf{Decoder 2} or \textbf{D2} in the paper.}
    \label{fig:decoder2}
  \end{subfigure}
  %\hfill
  \vspace{-0.9\baselineskip}
  \begin{subfigure}{0.48\linewidth}
   \includegraphics[width=\textwidth, trim={0 1cm 0 0},clip]{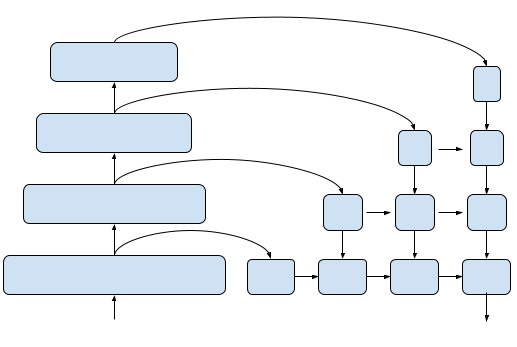}
    %\caption{Deep Layer Aggregation~\cite{yu2018deep}.}%(DLA). Refer to as \textbf{Decoder 3} or \textbf{D3} in the paper}
    \label{fig:decoder3}
  \end{subfigure}
  \hfill
  \begin{subfigure}{0.48\linewidth}
   \includegraphics[width=\textwidth, trim={0 1cm 0 0},clip]{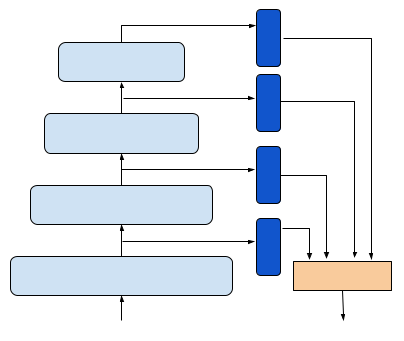}
    %\caption{Hierarchical intermediate map aggregation~\cite{bellitto2021hierarchical,saha2020recsal,wang2017deep}.}%. Refer to as \textbf{Decoder 1} or \textbf{D1} in the paper.}
    \label{fig:decoder4}
  \end{subfigure}
  %\hfill
  \caption{A taxonomy of decoding strategies commonly employed in video saliency prediction.
  Subfigures(top-left, top-right, bottom-left, bottom-right): Independent encoder and decoder, with no feature sharing between the two paths~\cite{djilali20203dsal,lai2019video,linardos2019simple,wu2020salsac,zou2021sta3d}; U-Net--like architecture, with features sharing between encoder and decoder~\cite{droste2020unified,jain2020vinet,li2021novel,min2019tased}; Deep Layer Aggregation~\cite{yu2018deep}; Hierarchical intermediate map aggregation~\cite{bellitto2021hierarchical,saha2020recsal,wang2017deep}.}
  \label{fig:decoders}
\end{figure}
As shown in the taxonomy presented in Fig.~\ref{fig:decoders}, a simpler class of approaches employs independent encoder and decoder, with no feature sharing between the two paths.
Among these, approaches based on recurrent layers typically model temporal dynamics at the bottleneck of the architecture~\cite{lai2019video,wu2020salsac,linardos2019simple}.
Non-recurrent architectures, instead, model time by means of 3D convolutions~\cite{zou2021sta3d,xie2018rethinking,djilali20203dsal},
Other approaches employ architectures similar to U-Net by introducing skip connections that encourage feature sharing between encoder and decoder. TASED~\cite{min2019tased} aggregates spatio-temporal features through the use of auxiliary pooling for reducing the temporal dimension. ViNet~\cite{jain2020vinet} integrates S3D features from multiple hierarchical levels by employing trilinear interpolation and 3D convolutions. UNISAL~\cite{droste2020unified} proposes a multi-objective unified framework for both 2D and 3D saliency with domain-specific modules and a lightweight recurrent architecture to handle temporal dynamics;
While single-decoder approaches are common, multi-decoder output integration has recently attracted interest. DVA~\cite{wang2017deep} and HD2S~\cite{bellitto2021hierarchical} %follow a similar approach, by extracting features at different abstraction levels and then each are fed to a independent decoder, each output of which are fused to produce the final prediction.
fuse maps predicted by independent decoders operating at different abstraction levels.
RecSal~\cite{saha2020recsal} predicts multiple saliency maps in a multi-objective training framework.
Recent works introduce more complex feature interactions among decoding paths, where high-resolution features are affected by deeper high-level features, as in DLA~\cite{yu2018deep} and TSFP-Net~\cite{chang2021temporal}. % includes multiple decoding paths that process a spatio-temporal feature pyramid. Since TSFP-Net shares aspects with hierarchical multi-decoder architectures, we employ DLA as a representative model for this family of architectures.
All of the approaches presented employ either a single-decoder architecture or a homogeneous multi-decoder one, where differences between decoders lie in the number of layers rather than in their structure. In our work, we propose an architecture which combines heterogeneous decoder structures, in order to better exploit their distinctive saliency prediction properties and thus increase computational efficiency.

\noindent\textbf{Knowledge distillation for visual saliency prediction.}
Knowledge distillation~\cite{hinton2015distilling,gou2021knowledge} is commonly employed to train an efficient \emph{student} model from a more complex \emph{teacher} model, with higher accuracy than when training the student directly from dataset labels.
Several knowledge distillation approaches have been recently proposed for video saliency prediction. SKD-DVA~\cite{li2019spatiotemporal} proposes spatio-temporal knowledge distillation with two teachers and two students, with each pair focusing on either spatial or temporal transfer.
SV2T-SS~\cite{zhang2019training} distills corresponding features of teacher and student (implemented as encoder-decoder networks), based on first- and second-order feature statistics transfer.
UVA-DVA~\cite{fu2020ultrafast} employs separate spatial and temporal teachers, whose knowledge is transferred to a single student model, which then fuses the resulting features in the final saliency prediction, achieving reasonable accuracy at impressive speed. Leveraging knowledge distillation for video salient object detection is the main theme of the work in~\cite{tang2020fast}.
The knowledge distillation setting proposed in our work differs from existing techniques in two main aspects: 1) we define a multi-objective distillation target on saliency maps directly; 2) we employ a hierarchical model as a teacher in order to further capture differences in saliency patterns extracted at multiple scales.

\section{Methodology}
\subsection{Overview}
\begin{figure*}
  \centering
   \includegraphics[width=0.95\textwidth]{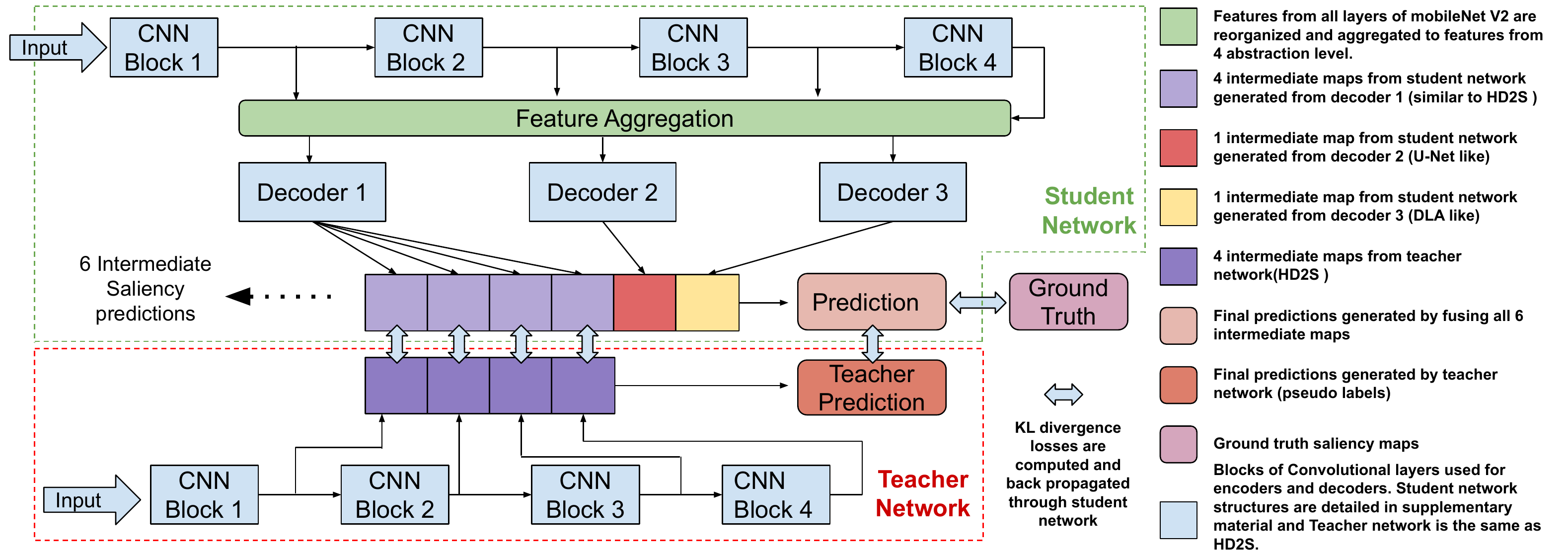}
  \caption{Overview of the proposed multi-decoder architecture with hierarchical knowledge distillation.}
  \label{fig:architecture}
\end{figure*}

The overall architecture of the proposed saliency prediction network with knowledge distillation is shown in Fig.~\ref{fig:architecture}. 
Following the taxonomy introduced in Sect.~\ref{sec:related_work}, a shared encoder extracts multi-level features that are then processed by three parallel decoding architectures: \textbf{decoder 1 (D1)} implements hierarchical intermediate maps aggregation (inspired by HD2S); \textbf{decoder 2 (D2)} employs a U-Net--like approach; \textbf{decoder 3 (D3)} is based on deep layer aggregation concepts (as in DLA~\cite{yu2018deep}).

The hierarchical aggregation decoder (i.e., \emph{decoder 1} in Fig.~\ref{fig:architecture}) produces four intermediate saliency maps from features extracted at different encoder layers; then, the set of predictions from all decoders are fused into the final prediction. At training time, we compute a supervised loss by comparing the final prediction to the ground-truth map, and a knowledge distillation loss on the final prediction and the intermediate maps extracted by D1 (all losses are based on Kullback-Leibler divergence between saliency maps; see Sect.~\ref{sec:losses}). In order to have a correspondence between intermediate maps produced by D1 and teacher maps, we employ HD2S as a teacher, since it naturally and semantically matches the decoder's hierarchical structure.

\subsection{Encoder structure}
Depthwise separable convolutions are widely used for efficient network design, as in MobileNetV2~\cite{sandler2018mobilenetv2}, commonly pre-trained on ImageNet and used as a backbone for lightweight models. In order to adapt it as a 3D video feature extractor, we follow the \emph{kernel inflation} approach introduced in~\cite{carreira2017quo} and already employed for static~\cite{hu2021fastsal} and dynamic~\cite{droste2020unified} saliency prediction.
A 2D convolutional kernel of size $C_\text{in} \times C_\text{out} \times H \times W$ can be inflated into a 3D kernel of size $C_{in} \times C_{out} \times T \times H \times W$ by replicating its weights along the temporal dimension $T$. This simple trick provides a convenient initialization that responds to common spatial patterns and can be gradually adapted to temporal dynamics during training, eliminating the burden of learning basic spatial structures from scratch.
Given the inflated MobileNetV2 encoder, we follow the approach in FastSal~\cite{hu2021fastsal} to extract four blocks of concatenated feature from the whole set of layers.

\subsubsection{Decoder structure and multiple prediction}
%The set of heterogeneous decoders, employed in our model, includes three types of decoding strategies, inspired to hierarchical map aggregation (\textbf{D1}), a U-Net--like architecture (\textbf{D2}) and deep layer aggregation (\textbf{D3}).
The set of heterogeneous decoders, employed in our model, includes \textbf{D1} (hierarchical map aggregation), \textbf{D2}(U-Net--like) and \textbf{D3} (deep layer aggregation). 
Our realizations of each of these approaches are designed to process the four input streams of features extracted by the encoder. D1 produces four intermediate saliency maps, while D2 and D3 produce a map each. The fusion layer that computes the final output map is implemented as a 1$\times$1 convolution of the predicted maps. Architectural details are reported in the supplementary materials.
As an additional efficiency consideration, we note that the high computation cost of many state-of-the-art approaches due to multiple-input/single-output (MISO) prediction, where a sequence of frames is used to predict a single saliency map, usually referring to the last frame. This provides a full context of previous frames to the model, but also means that, in order to predict $N$ saliency maps (without interpolation), $N$ forward passes are also required, with a proportional increase of computational power.
In order to further improve efficiency, we implement a multiple-input/multiple-output (MIMO) schema for output generation, by designing decoders that predict a number of saliency maps equal to the number of frames provided to the encoder. MIMO decoders can intrinsically make use of the similarity between consecutive saliency maps, and employ this information to reduce the computational power required to generate the same number of saliency maps by MISO decoders. Of course, the downside is that each frame has a different amount of surrounding context; however, in our experiments this has little impact on our model's accuracy.

\subsection{Knowledge distillation}
\label{sec:losses}
Given the presence of multiple decoders in our model, one of which also produces intermediate saliency maps, choosing a distillation approach to supervise the student's training is not trivial. As illustrated in Fig.~\ref{fig:architecture}, we carry out knowledge distillation by extracting intermediate and final outputs from a hierarchical teacher to supervise intermediate of one student decoder and final student outputs. 
%Our design of the distillation process is guided by several observations. First and foremost, it is necessary to provide a training signal at the very output of the model, in order to train the final fusion layer. Second, carrying out distillation at the representation level, by enforcing similarity between teacher and student features, defeats the purpose of having multiple decoders that are meant to recognize their own distinctive saliency patterns and should therefore be free to independently learn their own features. Also, feature-based supervision may require the use of adaptation layers to project teacher's features to the same dimension as the student's. Using saliency maps directly ensures that the output and target always have the same size. We therefore choose to use intermediate saliency maps from a hierarchical teacher, HD2S~\cite{bellitto2021hierarchical}, since this makes it possible in a natural way to affect the model at different depths of the encoder, without providing as strong a training signal as internal features.
Our design of the distillation process is guided by several observations. First and foremost, it is necessary to provide a training signal at the very output of the model, in order to train the final fusion layer. Second, carrying out distillation at the representation level, by enforcing similarity between teacher and student features, defeats the purpose of having multiple decoders that are meant to recognize their own distinctive saliency patterns and should therefore be free to independently learn their own features. 
Also, using saliency maps directly ensures that the output and target have the same size, so that the use of adaptation layers to match feature size of student's and teacher's can be avoided. We therefore choose to use intermediate saliency maps from a hierarchical teacher, HD2S~\cite{bellitto2021hierarchical}, since this makes it possible in a natural way to affect the model at different depths of the encoder, without providing as strong a training signal as internal features.

We formalize our knowledge distillation procedure as follows. Let $\vV$ be the space of video sequences and $\sS$ be the space of saliency maps (whether for the entire sequence or for a single frame); let $\mM$ be a family of models such that each element in $\mM$ is a function $M : \vV \rightarrow \sS^{n+1}$, which provides $n$ intermediate and one output saliency maps. We thus define a teacher $T \in \mM$ and student $S \in \mM$. For simplicity, the notations $S_i$ and $T_i$ will indicate the $i$-th map generated by, respectively, the student and teacher; indexes from 1 to $n$ will denote intermediate maps, while index $n+1$ will refer to the final output.
Saliency map distance is measured by Kullback-Leibler (KL) divergence:
\begin{equation}
\lL_\text{KL}\left(\xx, \yy\right) = \sum_{i} y_i \log \frac{y_i}{x_i},
\end{equation}
with $i$ iterating over spatial locations of the saliency maps.

At each training iteration, we sample a video sequence $\vv \in \vV$ and its ground-truth saliency $\sss \in \sS$.
%, and compute the teacher's predictions, to be used as pseudo-labels, $\left\{ \hat{\sss}_1, \dots, \hat{\sss}_{n+1} \right\} = T_\theta\left( \vv \right)$, and student's predictions $\left\{ \widetilde{\sss_1}, \dots, \widetilde{\sss}_{n+1} \right\} = S_\varphi\left( \vv \right)$.
The employed loss function aims at minimizing the KL divergence between student and teacher maps (both intermediate and final) and between (final) student and ground-truth maps:
\begin{equation}
%\begin{split}
\lL %\left(\widetilde{\sss}, \hat{\sss}, \sss\right)
=
\sum_{i=1}^{n+1} \lL_\text{KL}\big( S_i\left(\vv\right), T_i\left(\vv\right) \big) +
\lL_\text{KL} \big( S_{n+1}\left(\vv\right), \sss \big) .
%\end{split}
\end{equation}

\subsubsection{Training with auxiliary dataset}

The usage of unlabeled auxiliary datasets in a knowledge distillation setting has been shown to help boost performance~\cite{liao2005logistic,sattler2021fedaux,itahara2020distillation}. Following this approach, we introduce a new video distribution $\wW$, and extend the loss function with a term that measures the distance between student's predicted saliency maps and ``pseudo-labels'' (which are, in fact, also maps) provided by the teacher. As a result, given a pair of input videos $\vv \in \vV$ and $\ww \in \wW$, the new loss function becomes:
\begin{equation}
\begin{split}
\lL %\left(\widetilde{\sss}, \hat{\sss}, \sss\right)
=
\sum_{i=1}^{n+1} \lL_\text{KL}\big( S_i\left(\vv\right), T_i\left(\vv\right) \big) & +
\sum_{i=1}^{n+1} \lL_\text{KL}\big( S_i\left(\ww\right), T_i\left(\ww\right) \big) \\+
\lL_\text{KL} \big( & S_{n+1} \left(\vv\right), \sss \big).
\end{split}
\end{equation}

\subsubsection{Channel reduction with teacher assistant}
\label{sec:channel_reduction}
Previous works have shown that, with a suitable network design, it is possible to decrease the number of channels in the encoder's layers, in order to reduce the computational cost, without an excessive loss in accuracy~\cite{feichtenhofer2019slowfast}. %As one of the main objectives of our model, we investigate this idea.
Our channel reduction strategy applies multiple knowledge distillation iterations: at each of them, a new student is initialized by averaging the weights of each pair of consecutive kernels into a new kernel. Although kernel ordering is essentially random, this approach has been shown to provide a meaningful initialization to the new student. %~\cite{feichtenhofer2019slowfast}.
Additionally, we also explore the ``teacher assistant''~\cite{mirzadeh2020improved} distillation strategy: rather than using the original teacher to perform knowledge distillation on reduced-channel students, we employ the full-capacity student (i.e., before any channel reduction) as a new teacher.
As a result, by combining the channel reduction and teacher assistant, we encourage the model to distill more information while reducing computational cost.
%As a result, by combining the channel reduction and teacher assistant, we introduce a trade-off where we strengthen the training process, encouraging the model to further distill more informative features, while at the same time reducing the representational power and computational cost of the network.

\begin{table*}
  \setlength{\tabcolsep}{2.5pt}
  \centering
  %\footnotesize
  \caption{Comparison with SoA on the DHF1K and Hollywood2 test set in both the MISO and MIMO settings.}
  \label{tab:soa_dhf1k_hollywood}
  \begin{subtable}[t]{0.6\linewidth}
  \caption{Prediction accuracy and computational cost on the DHF1K test set. GMACs are estimated for 16 frames, hence the $\times$16 multiplication for MISO models. Models marked with a $^*$ are image saliency models.}
  \label{tab:soa_dhf1k}
  %\resizebox{0.9\linewidth}{!}{
  \begin{tabular}{@{}lcccccrr}
    \toprule
    \textbf{Models} & \textbf{AUC-J} & \textbf{SIM} & \textbf{sAUC} & \textbf{CC} & \textbf{NSS} & \textbf{GMACs} & ~~\textbf{\#params} \\%& input size\\
    \midrule
    \multicolumn{8}{c}{\emph{Multi-input/single-output (MISO) prediction}}\\
    \midrule
    SalGAN$^*$& 0.866 &0.262 &0.709	&0.370 &2.043 & 45.73$\times$16 & 31.92M \\%&1$\times$192$\times$256 \\
    FastSal$^*$& 0.887 &0.293 &0.712 &0.426 &2.330 & 2.64$\times$16 & 2.47M\\% & 1$\times$192$\times$256\\
    3DSal& 0.850 & 0.321 & 0.623 & 0.356 & 1.996 & 136.45$\times$16 & 46.15M \\%&6$\times$224$\times$224\\
    TASED & 0.895	&0.361	&0.712	&0.470	&2.667 & 91.75$\times$16 & 21.26M \\%&32$\times$224$\times$384\\
    ViNet& 0.908 &0.381	&\textbf{0.729}	&\textbf{0.511}	&2.872 & 115.28$\times$16 &31.1M\\% & 32$\times$224$\times$384 \\
    HD2S& 0.908	&\textbf{0.406}	&0.700	&0.503	&2.812 & 11.08$\times$16 & 29.8M \\%&16$\times$128$\times$192 \\
    \midrule
    TinyHD-S &\textbf{0.909} &0.396 &0.714 &0.505 &\textbf{2.921} & 5.57$\times$16 & 3.94M \\%& 16$\times$192$\times$256 \\
    %TinyHD-S(3T) &0.908 &0.405 &0.722 &\textbf{0.515} &\textbf{2.934} & 5.57$\times$16 & 3.94M \\%& 16$\times$192$\times$256 \\
    \bottomrule
        \multicolumn{8}{c}{\emph{Multi-input/multi-output (MIMO) prediction}}\\
    \toprule
    SalEMA& 0.890 &0.466 &0.667	&0.449	&2.574 & 640.16$\times$1 & 31.79M \\%&16$\times$192$\times$256 \\
    STRA-Net& 0.895	&0.355 &0.663 &0.458 &2.558 & 266.01$\times$3 & 168.02M \\%&5$\times$224$\times$224 \\
    UNISAL& 0.901 &\textbf{0.390} &0.691	&0.490	&2.776 & 19.42$\times$1 & 3.71M \\%&16$\times$224$\times$384 \\
    \midrule
    TinyHD-M &\textbf{0.905} &0.387 & \textbf{0.707}& \textbf{0.493} & \textbf{2.819} & 7.95$\times$1 & 3.92M\\% &16$\times$192$\times$256\\
    \bottomrule
  \end{tabular}
  %}
  \end{subtable}
  \hfill
  \begin{subtable}[t]{0.39\linewidth}
    \centering
    \caption{Prediction accuracy on Hollywood2}
    \label{tab:soa_hollywood}
    %\resizebox{\linewidth}{!}{% 
      \begin{tabular}{@{}lccccc}
        \toprule
     \textbf{Models} & \textbf{AUC-J} & \textbf{SIM} & \textbf{CC} & \textbf{NSS} \\
    \midrule
    \multicolumn{5}{c}{\emph{Multi-input/single-output prediction}}\\
    \midrule
    ACLNet &0.913 &0.757 &0.623	&3.086\\ 
    SalSAC &0.931 &0.529 &0.670 &3.356\\
    TASED &0.918 &0.507 &0.646 &3.302\\
    ViNet& 0.930 &0.550 &\textbf{0.693} &3.730\\
    HD2S& \textbf{0.936} &0.551 &0.670 &3.352\\
    \midrule
    TinyHD-S & 0.935 &\textbf{0.561} &0.690 &\textbf{3.815}\\
    \midrule
    \multicolumn{5}{c}{\emph{Multi-input/multi-output prediction}}\\
    \midrule
    SalEMA& 0.919 &0.487 &0.613 &3.186\\
    STRA-Net& 0.923	&0.536 &0.662 &3.478\\
    UNISAL& \textbf{0.934} &0.542 &0.673 &\textbf{3.901}\\
    \midrule
    TinyHD-M & \textbf{0.934} &\textbf{0.553} &\textbf{0.686} &3.744\\
    \bottomrule
    \end{tabular}
   %}%
  \end{subtable}%
  
\end{table*}

\begin{figure*}
	\centering
\begin{subfigure}[b]{0.11\textwidth}
	
	\includegraphics[width=\textwidth]{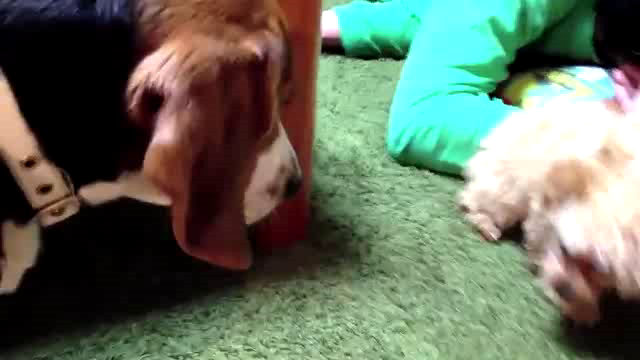}
	\label{example_0_0}
\end{subfigure}
\hspace{-0.5\baselineskip}
\begin{subfigure}[b]{0.11\textwidth}
	\includegraphics[width=\textwidth]{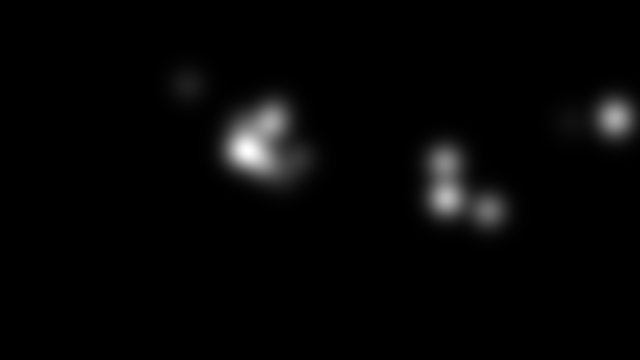}
	\label{example_0_1}
\end{subfigure}
\hspace{-0.5\baselineskip}
\begin{subfigure}[b]{0.11\textwidth}
	\includegraphics[width=\textwidth]{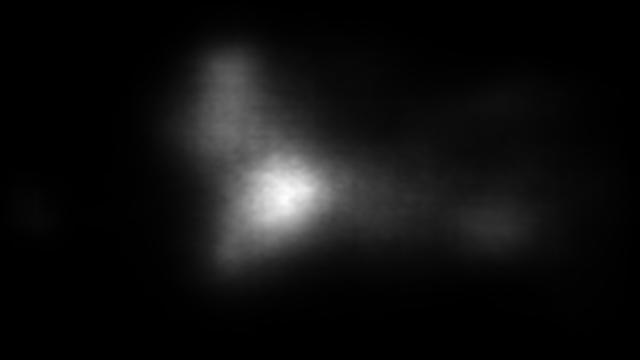}
	\label{example_0_8}
\end{subfigure}
\hspace{-0.5\baselineskip}
\begin{subfigure}[b]{0.11\textwidth}
	\includegraphics[width=\textwidth]{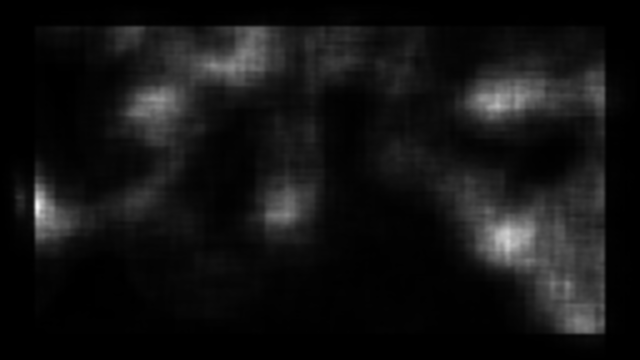}
	\label{example_0_2}
\end{subfigure}
\hspace{-0.5\baselineskip}
\begin{subfigure}[b]{0.11\textwidth}
	
	\includegraphics[width=\textwidth]{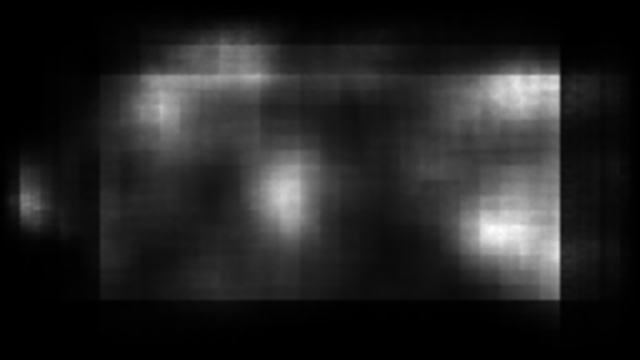}
	\label{example_0_3}
\end{subfigure}
\hspace{-0.5\baselineskip}
\begin{subfigure}[b]{0.11\textwidth}
	
	\includegraphics[width=\textwidth]{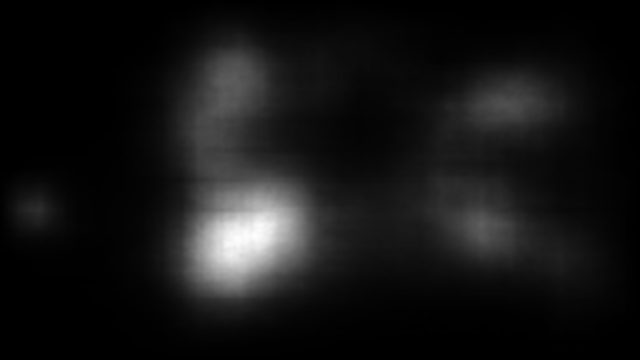}
	\label{example_0_4}
\end{subfigure}
\hspace{-0.5\baselineskip}
\begin{subfigure}[b]{0.11\textwidth}
	\includegraphics[width=\textwidth]{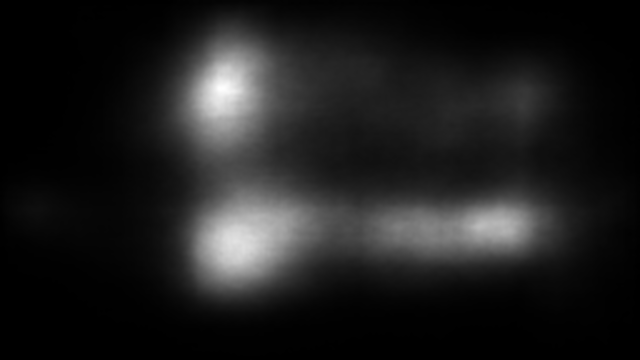}
	\label{example_0_5}
\end{subfigure}
\hspace{-0.5\baselineskip}
\begin{subfigure}[b]{0.11\textwidth}
	\includegraphics[width=\textwidth]{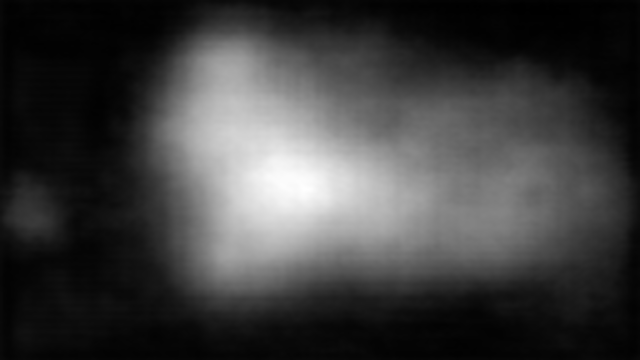}
	\label{example_0_6}
\end{subfigure}
\hspace{-0.5\baselineskip}
\begin{subfigure}[b]{0.11\textwidth}
	\includegraphics[width=\textwidth]{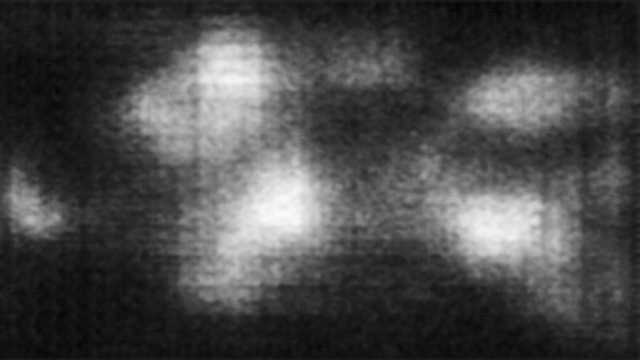}
	\label{example_0_7}
\end{subfigure}

\vspace{-1.0\baselineskip}
\begin{subfigure}[b]{0.11\textwidth}
	
	\includegraphics[width=\textwidth]{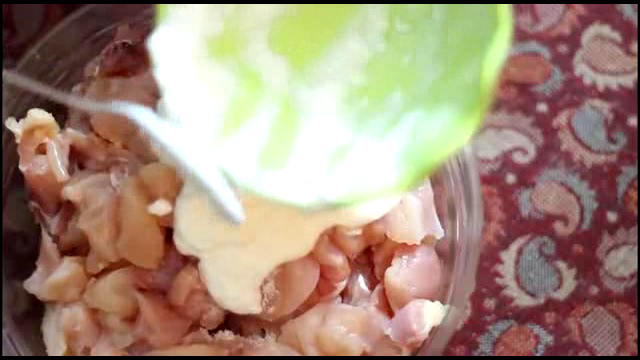}
	\label{example_1_0}
\end{subfigure}
\hspace{-0.5\baselineskip}
\begin{subfigure}[b]{0.11\textwidth}
	\includegraphics[width=\textwidth]{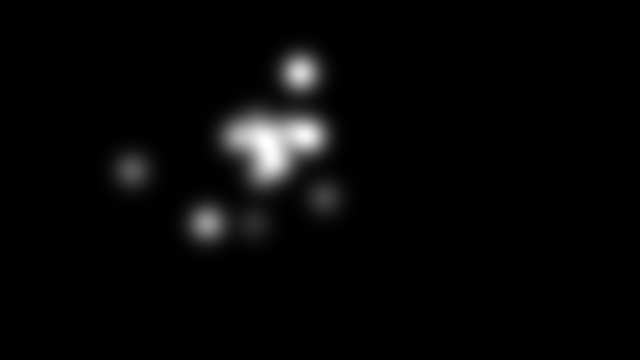}
	\label{example_1_1}
\end{subfigure}
\hspace{-0.5\baselineskip}
\begin{subfigure}[b]{0.11\textwidth}
	\includegraphics[width=\textwidth]{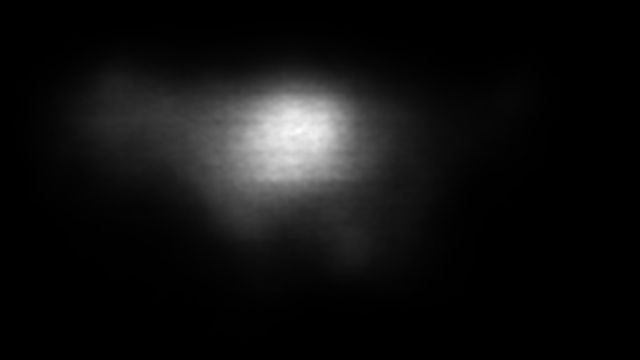}
	\label{example_1_8}
\end{subfigure}
\hspace{-0.5\baselineskip}
\begin{subfigure}[b]{0.11\textwidth}
	\includegraphics[width=\textwidth]{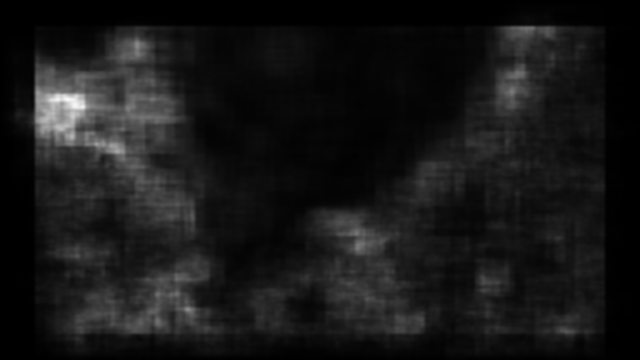}
	\label{example_1_2}
\end{subfigure}
\hspace{-0.5\baselineskip}
\begin{subfigure}[b]{0.11\textwidth}
	
	\includegraphics[width=\textwidth]{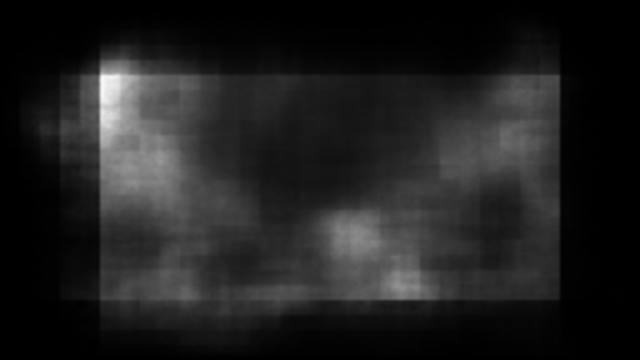}
	\label{example_1_3}
\end{subfigure}
\hspace{-0.5\baselineskip}
\begin{subfigure}[b]{0.11\textwidth}
	
	\includegraphics[width=\textwidth]{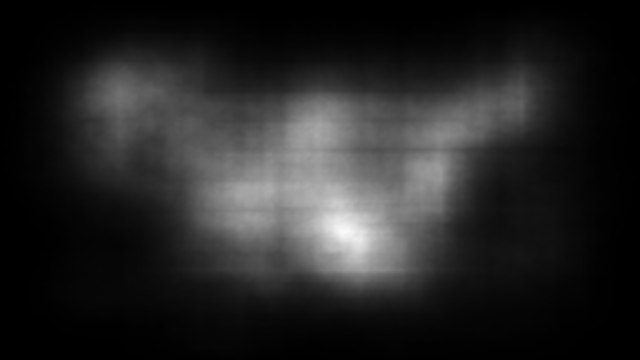}
	\label{example_1_4}
\end{subfigure}
\hspace{-0.5\baselineskip}
\begin{subfigure}[b]{0.11\textwidth}
	\includegraphics[width=\textwidth]{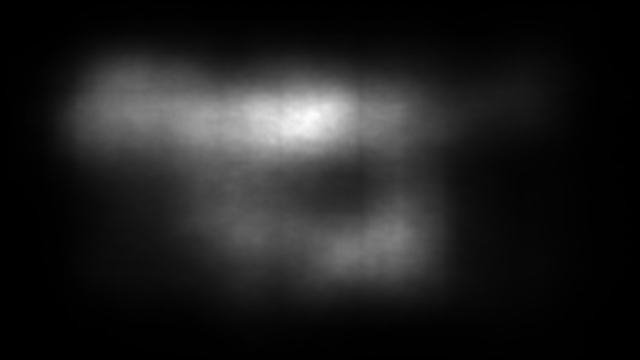}
	\label{example_1_5}
\end{subfigure}
\hspace{-0.5\baselineskip}
\begin{subfigure}[b]{0.11\textwidth}
	\includegraphics[width=\textwidth]{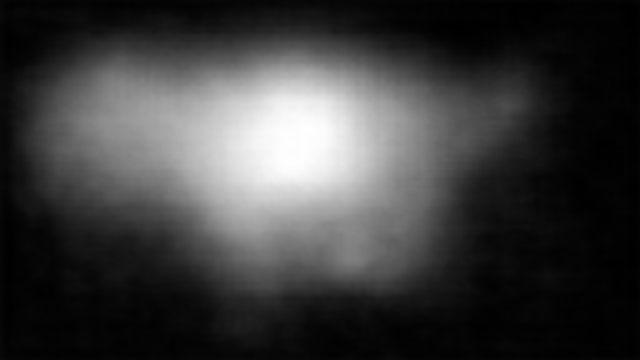}
	\label{example_1_6}
\end{subfigure}
\hspace{-0.5\baselineskip}
\begin{subfigure}[b]{0.11\textwidth}
	\includegraphics[width=\textwidth]{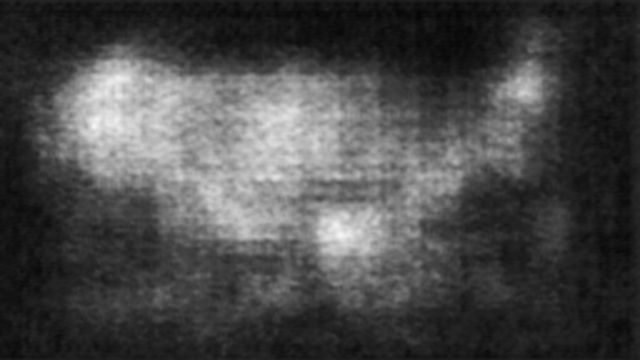}
	\label{example_1_7}
\end{subfigure}

\vspace{-1.0\baselineskip}
\begin{subfigure}[b]{0.11\textwidth}
	
	\includegraphics[width=\textwidth]{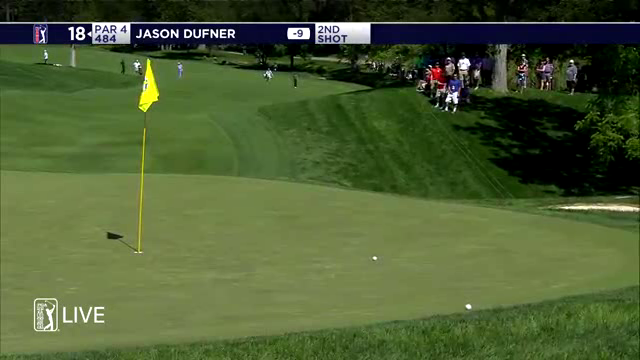}
	\label{example_2_0}
\end{subfigure}
\hspace{-0.5\baselineskip}
\begin{subfigure}[b]{0.11\textwidth}
	\includegraphics[width=\textwidth]{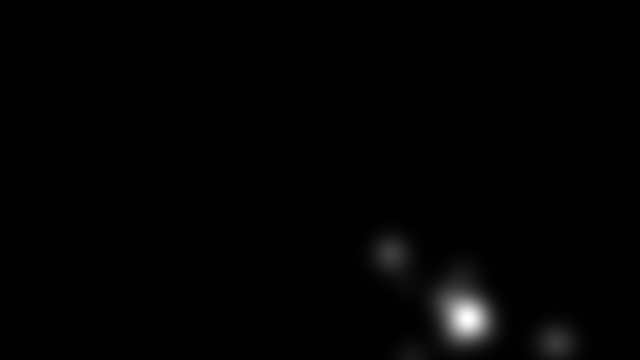}
	\label{example_2_1}
\end{subfigure}
\hspace{-0.5\baselineskip}
\begin{subfigure}[b]{0.11\textwidth}
	\includegraphics[width=\textwidth]{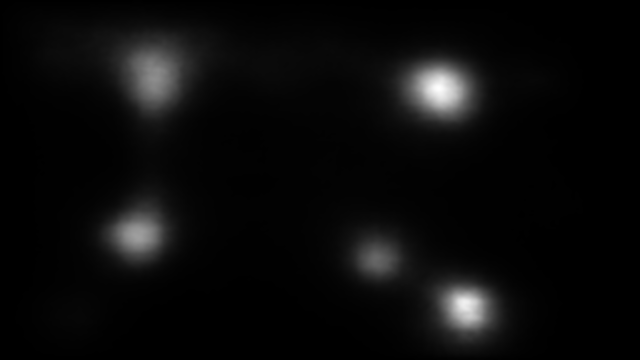}
	\label{example_2_8}
\end{subfigure}
\hspace{-0.5\baselineskip}
\begin{subfigure}[b]{0.11\textwidth}
	\includegraphics[width=\textwidth]{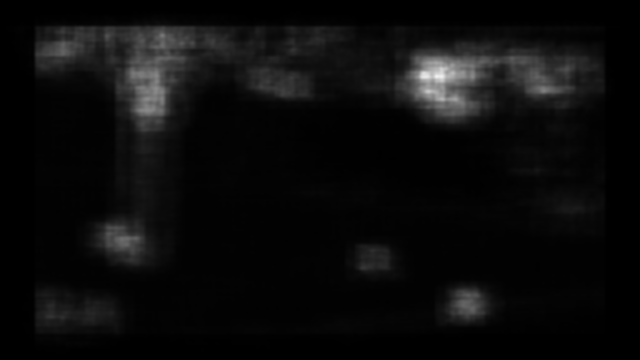}
	\label{example_2_2}
\end{subfigure}
\hspace{-0.5\baselineskip}
\begin{subfigure}[b]{0.11\textwidth}
	
	\includegraphics[width=\textwidth]{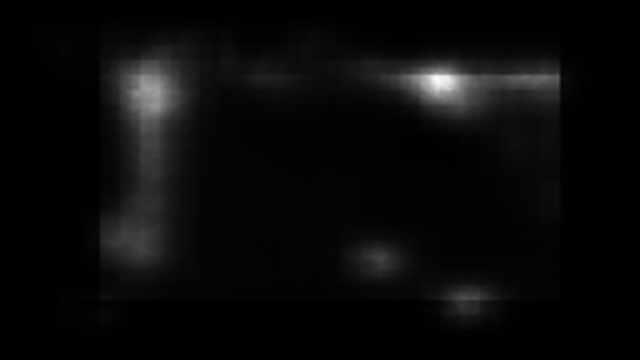}
	\label{example_2_3}
\end{subfigure}
\hspace{-0.5\baselineskip}
\begin{subfigure}[b]{0.11\textwidth}
	
	\includegraphics[width=\textwidth]{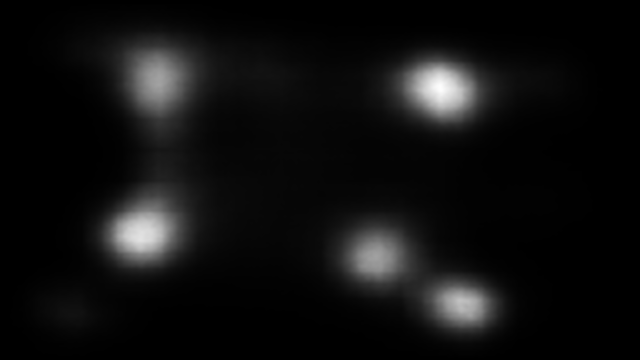}
	\label{example_2_4}
\end{subfigure}
\hspace{-0.5\baselineskip}
\begin{subfigure}[b]{0.11\textwidth}
	\includegraphics[width=\textwidth]{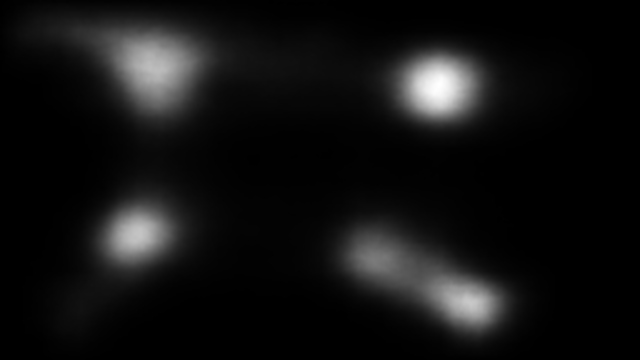}
	\label{example_2_5}
\end{subfigure}
\hspace{-0.5\baselineskip}
\begin{subfigure}[b]{0.11\textwidth}
	\includegraphics[width=\textwidth]{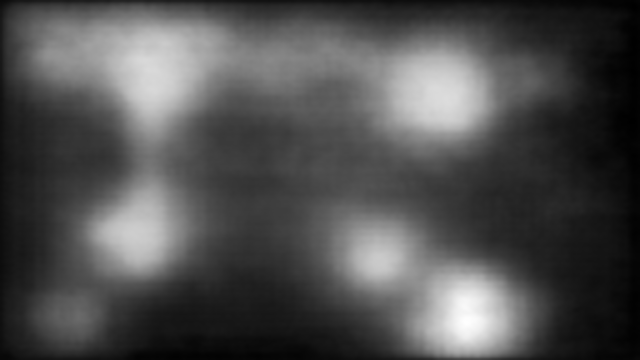}
	\label{example_2_6}
\end{subfigure}
\hspace{-0.5\baselineskip}
\begin{subfigure}[b]{0.11\textwidth}
	\includegraphics[width=\textwidth]{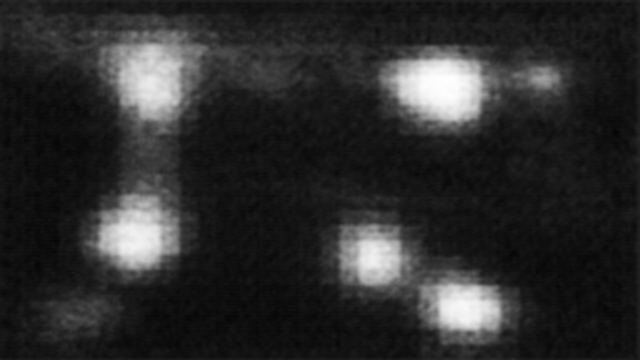}
	\label{example_2_7}
\end{subfigure}

\vspace{-1.0\baselineskip}
\begin{subfigure}[b]{0.11\textwidth}
	
	\includegraphics[width=\textwidth]{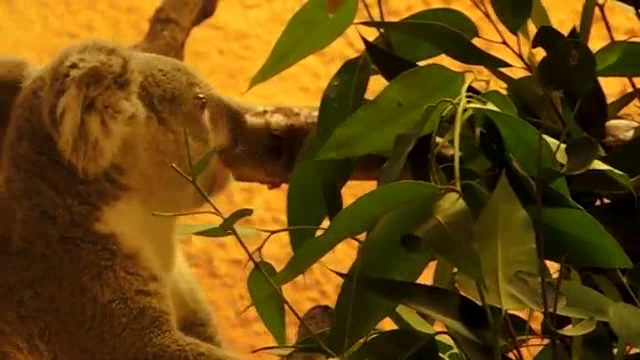}
	\caption*{Frame}
	\label{example_3_0}
\end{subfigure}
\hspace{-0.5\baselineskip}
\begin{subfigure}[b]{0.11\textwidth}
	
	\includegraphics[width=\textwidth]{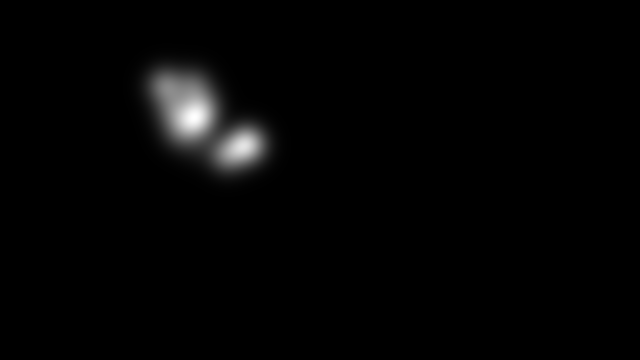}
	\caption*{GT}
	\label{example_3_1}
\end{subfigure}
\hspace{-0.5\baselineskip}
\begin{subfigure}[b]{0.11\textwidth}
	\includegraphics[width=\textwidth]{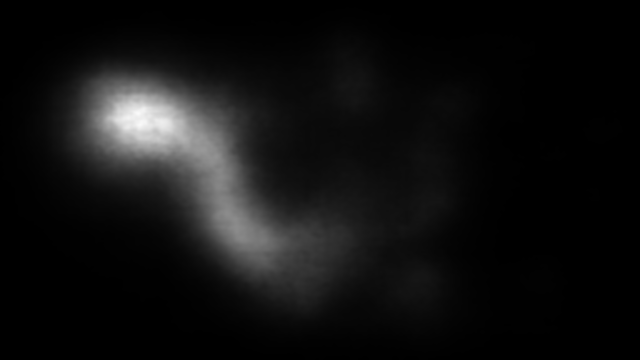}
	\caption*{Output}
	\label{example_3_8}
\end{subfigure}
\hspace{-0.5\baselineskip}
\begin{subfigure}[b]{0.11\textwidth}
	
	\includegraphics[width=\textwidth]{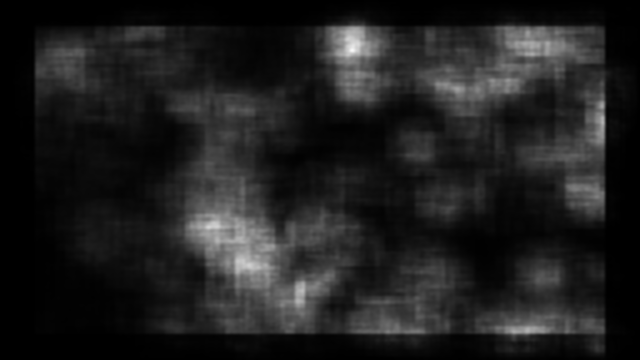}
	\caption*{D1 (1)}
	\label{example_3_2}
\end{subfigure}
\hspace{-0.5\baselineskip}
\begin{subfigure}[b]{0.11\textwidth}
	
	\includegraphics[width=\textwidth]{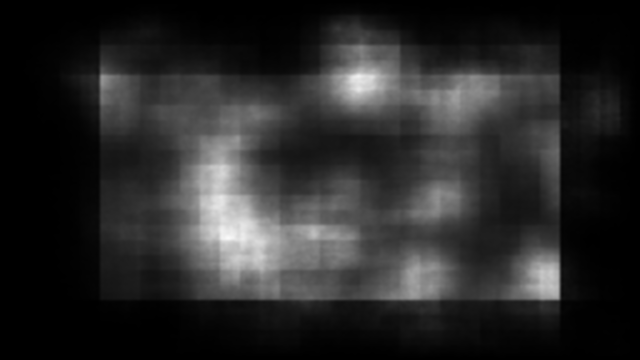}
	\caption*{D1 (2)}
	\label{example_3_3}
\end{subfigure}
\hspace{-0.5\baselineskip}
\begin{subfigure}[b]{0.11\textwidth}
	
	\includegraphics[width=\textwidth]{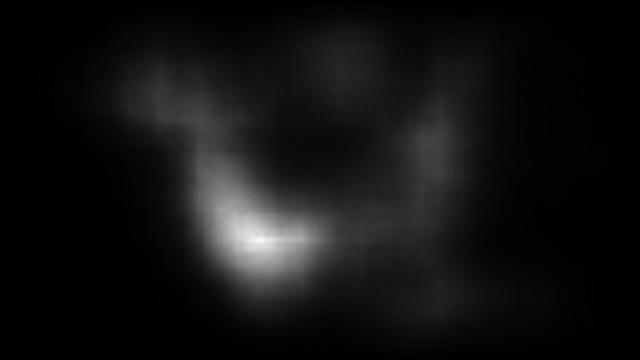}
	\caption*{D1 (3)}
	\label{example_3_4}
\end{subfigure}
\hspace{-0.5\baselineskip}
\begin{subfigure}[b]{0.11\textwidth}
	\includegraphics[width=\textwidth]{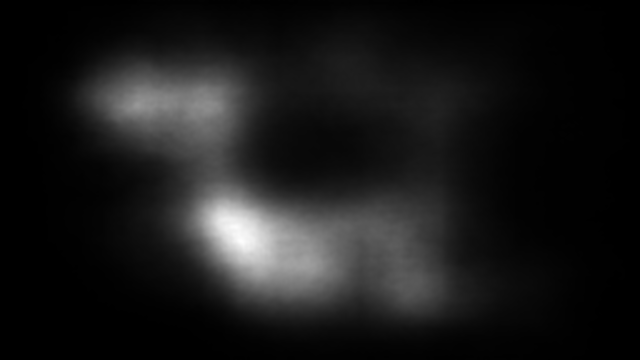}
	\caption*{D1 (4)}
	\label{example_3_5}
\end{subfigure}
\hspace{-0.5\baselineskip}
\begin{subfigure}[b]{0.11\textwidth}
	\includegraphics[width=\textwidth]{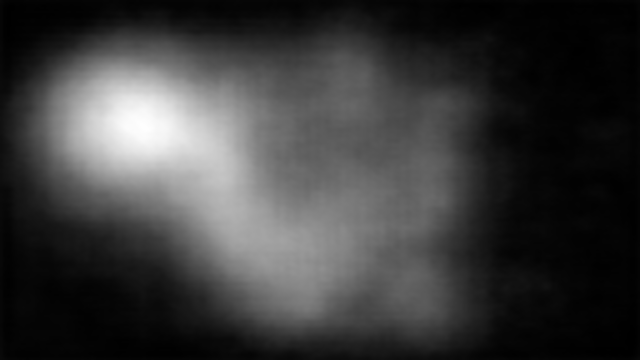}
	\caption*{D2}
	\label{example_3_6}
\end{subfigure}
\hspace{-0.5\baselineskip}
\begin{subfigure}[b]{0.11\textwidth}
	\includegraphics[width=\textwidth]{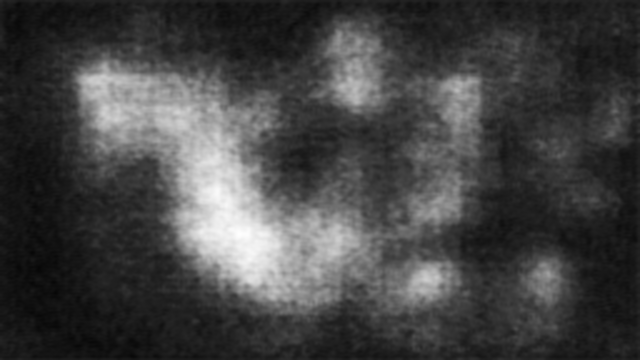}
	\caption*{D3}
	\label{example_3_7}
\end{subfigure}
	\caption{Examples of video saliency maps predicted by the proposed model, as well as intermediate maps by multiple decoders. Values between parentheses indicate one of the intermediate saliency maps by decoder D1.}
	\label{fig:saliency_example}
\end{figure*}
\section{Experiments}
\subsection{Datasets and Metrics}
We conduct experiments on DHF1K~\cite{wang2019revisiting}, UCF-Sports~\cite{rodriguez2008action,mathe2014actions} and Hollywood2~\cite{marszalek2009actions,mathe2014actions} datasets, commonly employed to evaluate video saliency prediction.

\textbf{DHF1K} contains 1000 videos split into 600/100/300 for training, validation, and test (unreleased). Eye fixations are collected from 17 participants in free-viewing experiments.
\textbf{UCF-Sports} is a task-driven dataset that includes 150 videos (103 for training, 47 for test) covering 9 sport activities. Participants were asked to identify the activity in each video sequence.
\textbf{Hollywood2} includes 1707 videos extracted from 69 movies and categorized between 12 action classes. At data collection, 3 observers are free-viewing, 12 observers are asked recognize the action, and 4 observers are asked to recognize the scene. 823 videos are used for training and 884 for test.
%\begin{itemize}
%    \item \textbf{DHF1K} contains 1000 videos split into 600/100/300 for training, validation, and test (unreleased). Eye fixations are collected from 17 participants using an eye tracking device in free-viewing experiments.
%    \item \textbf{UCF-Sports} is a task-driven dataset that includes 150 videos (103 for training, 47 for test) covering 9 sport activities. Participants were asked to identify the activity in each video sequence.
%    \item \textbf{Hollywood2} includes 1707 videos extracted from 69 movies and categorized between 12 action classes. At data collection, 3 observers are free-viewing, 12 observers are asked recognize the action, and 4 observers are asked to recognize the scene. 823 videos are used for training and 884 for test.
%\end{itemize}
We also employ the Kinetic-400~\cite{kay2017kinetics} action recognition benchmark as auxiliary dataset, used by the teacher to generate additional training inputs with pseudo-labels.
For evaluation purposes, we report results in terms of the standard metrics for video saliency prediction~\cite{wang2017deep}: AUC-Judd (AUC-J), AUC-Borji (AUC-B), Linear Correlation Coefficient (CC), Normalized Scanpath Saliency (NSS), and Similarity Metric (SIM). %Shuffled AUC (sAUC) is reported in DHF1K testing set.
\subsection{Training procedure}
%To construct an encoder with inflated 3D CNN from MobileNet V2, we use pretrained weights from ImageNet. All decoders are also lightweight and use, as a basic building block, an inverted residual module, similarly to MobileNet V2.
%During each training epoch, one video clip is uniformly sampled from DHF1K and one video clip uniformly sampled from the full Kinetic400 dataset. In the experiments dealing only with DHF1K, two video clips are uniformly sampled from a video. When only Kinetics-400 is used, $600 \times 2$ clips are  sampled from the full dataset in order to match the number of clips used in DHF1K. When training our model with UCF-sport and Kinetic-400, for every clip sampled in UCF-sport, we sample 10 clips from Kinetics-400, due to the limited size of UCF-sport dataset.
Models are trained for 200 epochs using mini-batch stochastic gradient descent, with a mini-batch size of 12. The initial learning rate is 0.01, and it is reduced by a factor of 0.1 at epochs 100, 150, and 180. Input sequence length is 16 frames, spatially resized to 192$\times$256. We carry out data augmentation by means of random horizontal flips; in our experiments, spatial resize and cropping do not lead to significant benefits.
When the teacher assistant strategy is employed for channel reduction, we perform two additional knowledge distillation, each time training a new student network whose encoder contains, respectively, half and a quarter of the original number of channel at each encoder layer.

\subsection{Performance comparison with state-of-the-art models}
%\todo{Moved: We refer the model that generates one frame at a time as \textbf{TinyHD-S} and the model that generates the multiple frames as \textbf{TinyHD-M}. In fact, to generate 16 saliency maps, Table~\ref{tab:soa_dhf1k} shows that the M version uses about 10 times less computational power than the S version while using similar number of parameters.}
In these experiments, we report results of our model in both the MISO and MIMO configurations (respectively, \textbf{TinyHD-S} and \textbf{TinyHD-M}), trained with the auxiliary unlabeled dataset but \emph{without} channel reduction that using the teacher assistant strategy (which introduces trade-offs between accuracy and computational costs that will be discussed later).
%We compare our model to state-of-the-art methods on the test sets for DHF1K, UCF-Sports and Hollywood2 (all results are from the DHF1K leaderboard).% except for 3DSal~\cite{djilali20203dsal}).
We also report the number of multiply-accumulate operations (MAC) carried out by each method\footnote{Values are computed from official implementations when available and from our own implementations otherwise.} to generate a 16-frame saliency sequence. 
Results on DHF1K are shown in Table~\ref{tab:soa_dhf1k}. In the MISO configuration, our model is on par with state-of-the-art methods (and even better on NSS), but only employs a fraction of the their computational cost. In the MIMO configuration, our method sets a new state of the art, outperforming (on four metrics out of five) also UNISAL, which has a similar number of parameters but is about twice as demanding in terms of GMACs. %Furthermore, it can be observed that the MIMO-variant of our approach requires 10 times less computational power than the S counterpart while using a similar number of parameters.
Fig.~\ref{fig:saliency_example} presents a few examples of saliency predictions by our model\footnote{More examples are provided in the supplementary materials, as well as a visual comparison with state-of-the-art models.}. For each example, we also show the intermediate maps provided by each decoder. 
%For short, we identify as \textbf{D1} the decoder that implements hierarchical intermediate maps aggregation (inspired by HD2S), whose output is a set of four maps at different feature abstraction levels; \textbf{D2} employs a U-Net--like approach (similarly to TASED-net~\cite{min2019tased} and ViNet~\cite{jain2020vinet}; \textbf{D3} is based on deep layer aggregation (as in DLA~\cite{yu2018deep}). 
Qualitatively, our model predicts reasonable saliency regions, sometimes identifying additional elements not included in the ground truth (e.g., the third example). Intermediate maps also exhibit a certain variability, although similar patterns can be found in pairs (e.g., maps 1-2 and maps 3-4 from D1, and maps from D2 and D3). In general, the highest-level map from D1 (the fourth) mostly affects the output prediction: this is expected, since the corresponding architecture matches the teacher's. However, the fusion layer includes all information from intermediate maps, as shown in the last example, where two salient areas identified by the highest-level map from D1 are discarded.

Table~\ref{tab:soa_hollywood} and ~\ref{tab:ucf_soa} report results on Hollywood2 and UCF-Sports. While the model performs very well on the former, especially in the more efficient MIMO setting, ViNet and UNISAL achieve higher accuracy on UCF-Sports. This may be due to the lower performance of the HD2S teacher on that specific dataset, and to the arguable suitability of UCF-Sports as a video saliency prediction benchmark: the vast majority of its videos has fewer than 100 frames, and user fixations are driven by action classification, rather than free-viewing saliency~\cite{bellitto2021hierarchical}.

\begin{table}[t]
\setlength{\tabcolsep}{2.0pt}
  \centering
  \caption{Performance comparison on UCF-Sports in both the MISO and MIMO settings.}
  \label{tab:soa_ucf_hollywood}
  %\begin{subtable}[t]{0.99\linewidth}
  %  \centering
  %  \caption{Hollywood2}
  %  \label{tab:hollywood_soa}
    %\resizebox{\linewidth}{!}{% 
  %    \begin{tabular}{@{}lccccc}
  %      \toprule
  %   \textbf{Models} & \textbf{AUC-J} & \textbf{SIM} & \textbf{CC} & \textbf{NSS} \\
  %  \midrule
  %  \multicolumn{5}{c}{\emph{Multi-input/single-output prediction}}\\
  %  \midrule
  %  ACLNet &0.913 &0.757 &0.623	&3.086\\ 
  %  SalSAC &0.931 &0.529 &0.670 &3.356\\
  %  TASED &0.918 &0.507 &0.646 &3.302\\
  %  ViNet& 0.930 &0.550 &\textbf{0.693} &3.730\\
  %  HD2S& \textbf{0.936} &0.551 &0.670 &3.352\\
  %  \midrule
  %  TinyHD-S & 0.935 &\textbf{0.561} &0.690 &\textbf{3.815}\\
  %  \midrule
  %  \multicolumn{5}{c}{\emph{Multi-input/multi-output prediction}}\\
  %  \midrule
  %  SalEMA& 0.919 &0.487 &0.613 &3.186\\
  %  STRA-Net& 0.923	&0.536 &0.662 &3.478\\
  %  UNISAL& \textbf{0.934} &0.542 &0.673 &\textbf{3.901}\\
  %  \midrule
  %  TinyHD-M & \textbf{0.934} &\textbf{0.553} &\textbf{0.686} &3.744\\
  %  \bottomrule
  %  \end{tabular}
   %}%
  %\end{subtable}%
  %\vfill
  %\begin{subtable}[t]{0.99\linewidth}
    \centering
    %\caption{UCF-Sports}
    \label{tab:ucf_soa}
    %\resizebox{\linewidth}{!}{% 
      \begin{tabular}{@{}lccccc}
        \toprule
     \textbf{Models} & \textbf{AUC-J} & \textbf{SIM} & \textbf{CC} & \textbf{NSS} \\
    \midrule
    \multicolumn{5}{c}{\emph{Multi-input/single-output prediction}}\\
    \midrule
    ACLNet &0.897&0.406 &0.510 & 2.567\\ 
    3DSal& 0.881 & 0.478 & 0.590 & 2.802 \\
    TASED & 0.899	& 0.469 &0.582	&2.920 \\
    ViNet& \textbf{0.924} & \textbf{0.522} &\textbf{0.673}	& \textbf{3.620}\\
    HD2S& 0.904	& 0.507 &0.604	&3.114 \\
    \midrule
    %TinyHD-S(V) &\textbf{0.925} &0.516  &0.650 &3.429 \\  
    TinyHD-S & 0.918 &0.510 &0.624 &3.280\\
    \midrule
    \multicolumn{5}{c}{\emph{Multi-input/multi-output prediction}}\\
    \midrule
    SalEMA& 0.906 & 0.431 	&0.544	&2.638 \\
    STRA-Net& 0.910	&0.479  &0.593 &3.018 \\
    UNISAL& \textbf{0.918} &  \textbf{0.523} & \textbf{0.644} &\textbf{3.381}\\
    \midrule
    %TinyHD-M(V) &\textbf{0.925} &\textbf{0.529} &\textbf{0.655} & \textbf{3.452} \\
    TinyHD-M & 0.911  &0.499 &0.609 &3.234\\
    \bottomrule
    \end{tabular}
   %}%
   
  %\end{subtable}%
\end{table}

\subsection{Ablation studies}
\begin{table}[t]
\setlength{\tabcolsep}{1.4pt}
%\begin{subtable}[t]{0.99\linewidth}
%\centering
%\begin{tabular}{@{}lccc|ccc}
%\toprule
%& \multicolumn{3}{c}{\emph{GMACs}} & \multicolumn{3}{c}{\emph{\#params}}\\
%Decoder & $\times 1$ & $\times 2$ & $\times 3$ & $\times 1$ & $\times 2$ & $\times 3$\\
%\midrule
%D1& 3.55$\times 16$ & 2.88$\times 16$ & 2.45$\times 16$ & 2.55M & 3.57M & 2.53M\\
%D2& 5.45$\times 16$ & 4.11$\times 16$ & 3.24$\times 16$ & 2.75M & 4.78M & 2.70M\\
%D3& 7.35$\times 16$ & 5.33$\times 16$ & 4.03$\times 16$ & 2.95M & 6.00M & 2.88M\\
%\midrule
%TinyHD-S & \multicolumn{3}{c}{5.57$\times 16$} & \multicolumn{3}{c}{3.94M}\\
%\bottomrule
%\end{tabular}
%\end{subtable}
%\begin{subtable}[t]{0.99\linewidth}
\caption{Performance of our architecture with homogeneous decoders, on the DHF1K validation set. Number of parameters of models with homogeneous decoders are: D1$\times 1$ (2.55M), D2$\times 1$ (3.57M). D3$\times 1$ (2.53M); D1$\times 2$ (2.75M), D2$\times 2$ (4.78M). D3$\times 2$ (2.70M); D1$\times 3$ (2.95M), D2$\times 3$ (6.00M). D3$\times 3$ (2.88M); TinyHD-S (3.94M).}
\label{tab:decoder_single}
\centering
  %\begin{subtable}[t]{0.8\linewidth}
    %\centering
    %\caption{Experiments using identical stacked decoders.}
    %\label{tab:single_stack}
   % \resizebox{\linewidth}{!}{% 
      \begin{tabular}{@{}lcccccc}
      %\centering
      \toprule
      \textbf{Decoder} & \textbf{AUC-J} & \textbf{AUC-B} & \textbf{CC} & \textbf{NSS} & \textbf{SIM} & \textbf{GMACs} \\
      \midrule
      D1$\times 1$ &0.8993 &0.8210 &0.4881 &2.8163 &\textbf{0.3939}& 3.55$\times$16 \\
      D2$\times 1$ &0.9040 &0.8235 &0.4837 &2.7976 &0.3820& 2.88$\times$16 \\
      D3$\times 1$ &0.9034 &0.8248 &0.4836 &2.7851 &0.3794& 2.45$\times$16 \\
      \midrule
      D1$\times 2$ &0.8998 & 0.8195 &0.4882 &2.8256 &0.3928 & 5.45$\times$16 \\
      D2$\times 2$ &0.9046 & 0.8251 &0.4855 &2.8117 &0.3806 & 4.11$\times$16 \\
      D3$\times 2$ &0.9046 & 0.8239 &0.4864 &2.8095 &0.3819 & 3.24$\times$16 \\
      \midrule
      D1$\times 3$ &0.9013 &0.8253 &0.4922 &2.8420 &0.3924 & 7.35$\times$16 \\
      D2$\times 3$ &0.9049 &\textbf{0.8266} &0.4847 &2.8042 &0.3774 & 5.33$\times$16 \\
      D3$\times 3$ &0.9047 &0.8242 &0.4845 &2.7967 &0.3799 & 4.03$\times$16 \\
      \midrule
      TinyHD-S &\textbf{0.9075} &0.8244 &\textbf{0.4945} &\textbf{2.8735} &0.3887& 5.57$\times$16 \\
      \bottomrule
      \end{tabular}
%\end{subtable}
\end{table}
In order to experimentally substantiate our architectural and methodological choices, we carry out a set of ablation studies on each component of the model. the results of these experiments are reported on the DHF1K validation set, since testing set is not publicly available. %These experiments are carried out by splitting the DHF1K training set into a smaller training set and validation set (with 90/10 proportion), and using the DHF1K validation set as a test set.
First, we assess the effect of our heterogeneous multi-decoder strategy, evaluating the model's performance under several decoder configurations. We carry out this experiment in the MISO configuration, which achieves higher accuracy, as shown in Table~\ref{tab:soa_dhf1k}. In order to demonstrate the importance of combining different decoder architectures, Table~\ref{tab:decoder_single} reports results when using homogeneous decoders in our architecture. Table~\ref{tab:decoder_single} show that the heterogeneous approach generally performs better than configurations with a single decoder type, most remarkably in the NSS metric.
For the sake of completeness, we also show configurations where a smaller number of homogeneous decoders are employed; these setups are, of course, more computationally efficient, but exhibit lower performance on average in the accuracy metrics.

In the second part of our ablation study, we evaluate of the impact of our knowledge distillation strategy. Table~\ref{tab:kd_components} reports the results obtained by the proposed model, in MISO configuration, when trained on ground-truth maps only, and when gradually adding knowledge distillation terms on DHF1K and on Kinetics-400, using HD2S as teacher. The full loss setting achieves better performance on average --- as previously. This is most evident in the NSS metric.

\begin{table}
\setlength{\tabcolsep}{2.0pt}
  \centering
  \caption{Impact of loss terms on our model in the MISO configuration, starting from training on ground-truth (GT) maps only, and gradually adding knowledge distillation terms on DHF1K (target dataset or TD) and on Kinetics-400 (auxiliary dataset or AD), using HD2S as a teacher.}
  \label{tab:kd_components}
  \begin{tabular}{@{}lccccc}
    \toprule
    \textbf{Loss term} & \textbf{AUC-J} & \textbf{AUC-B} & \textbf{CC} & \textbf{NSS} & \textbf{SIM} \\
    \midrule
    %Ground-truth maps & 0.9033 & \textbf{0.8286} &0.4864 &2.7680 & 0.3765\\
    %& \hspace{0.2cm} + teacher & 0.9033 &0.8286 &0.4864 &2.7680 & 0.3765\\
    %\hspace{0.2cm} + K.D. on target dataset & 0.9058 &0.8237 &0.4875 &2.8182 &0.3846\\
    %\hspace{0.4cm} + K.D. on auxiliary dataset & \textbf{0.9075} & 0.8244 & \textbf{0.4945} & \textbf{2.8735} & \textbf{0.3887}\\
    GT maps & 0.9033 & \textbf{0.8286} &0.4864 &2.7680 & 0.3765\\
    \hspace{0.2cm} + K.D. on TD & 0.9058 &0.8237 &0.4875 &2.8182 &0.3846\\
    \hspace{0.4cm} + K.D. on AD & \textbf{0.9075} & 0.8244 & \textbf{0.4945} & \textbf{2.8735} & \textbf{0.3887}\\
    %& Teacher only &0.8943 &0.8018 &0.4527 &2.6486 &0.3645\\
    %& Teacher \& label & 0.9058 &0.8237 &0.4875 &2.8182 &0.3846\\
    \bottomrule
  \end{tabular}
\end{table}

\subsection{Channel reduction with teacher assistant}
\label{sec:effect_channel_reduction}
Finally, we investigate further reducing computational costs by means of our channel reduction strategy: multiple distillation steps are carried out, with each student progressively halving its number of encoding and decoding features, as described in Sect.~\ref{sec:channel_reduction}. We also evaluate the performance of this approach when training on the original teacher (HD2S) and when using the ``teacher assistant'' technique, with the full-capacity student used as a teacher.
Table~\ref{tab:reduce_channel_dhf1k_val} reports results, on both MISO and MIMO settings, after one and two reduction steps steps, respectively resulting in models with half (marked as $\times\frac{1}{2}$) and a quarter (marked as $\times\frac{1}{4}$) of the original number of convolutional features (marked as $\times 1$). Rows with ``+TA'' denote the use of the full-capacity student as teacher for knowledge distillation, rather than HD2S.
As expected, channel reduction introduces a trade-off between retaining the accuracy of the original model and reducing computational costs. As multiply-accumulate operations and model parameters are significantly reduced, accuracy also decreases, most evidently in the NSS and, to a smaller extent, in the SIM metrics. It is noteworthy that configurations employing a teacher assistant outperform the counterpart using HD2S. 

\begin{table}
\setlength{\tabcolsep}{2.0pt}
  \centering
  \caption{Performance of the proposed model when employing channel reduction and teacher assistant distillation.}
  \label{tab:reduce_channel_dhf1k_val}

  \begin{subtable}[t]{0.99\linewidth}
    \caption{Number of parameters of models with reduced channels and GMACs reported on generating 16 output saliency maps.}
    \centering
    \begin{tabular}{@{}lccc|ccc}
    \toprule
    & \multicolumn{3}{c}{\textbf{GMACs}} & \multicolumn{3}{c}{\textbf{\#params}}\\
    \textbf{Models} & $\times 1$ & $\times \frac{1}{2}$ & $\times \frac{1}{4}$ & $\times 1$ & $\times \frac{1}{2}$ & $\times \frac{1}{4}$\\
    \midrule
    TinyHD-S& 89.12 & 59.52 & 37.44 & 3.94M & 1.37M & 513.1k\\
    TinyHD-M& 7.95 & 6.92 & 4.06 & 3.92M & 1.37M & 515.3k\\
    \bottomrule
    \end{tabular}
    \end{subtable}
  
  \begin{subtable}[t]{0.99\linewidth}
  \centering
  \caption{Performance of channel reduction reported on DHF1K validation set in both the MISO and MIMO settings.}
  \begin{tabular}{@{}lccccc}
    \toprule
    \textbf{Models} & \textbf{AUC-J} & \textbf{AUC-B} & \textbf{CC} & \textbf{NSS} & \textbf{SIM}\\
    \midrule
    \multicolumn{6}{c}{\emph{Multi-input/single-output prediction}}\\
    \midrule
    %TinyHD-S$\times 1$ &0.9074 &0.8243 & 0.4944 & 2.8725 & &5.57Gx16 & 3.94M \\
    TinyHD-S$\times 1$& \textbf{0.9075} & 0.8244 & \textbf{0.4945} & \textbf{2.8735} & \textbf{0.3887} \\
    \midrule
    TinyHD-S$\times \frac{1}{2}$ &0.9038 &\textbf{0.8331} &0.4754 &2.7194 &0.3641\\
    \hspace{0.2cm} +TA & 0.9052 &0.8330 &0.4805 &2.7317 &0.3684\\
    \midrule
    TinyHD-S$\times \frac{1}{4}$ &0.9005 &0.8285 &0.4560 &2.5830 &0.3514\\
    \hspace{0.2cm} +TA & 0.9018 & 0.8318 & 0.4667 & 2.6329 & 0.3569\\
    \midrule
    \multicolumn{6}{c}{\emph{Multi-input/multi-output prediction}}\\
    \midrule
    TinyHD-M$\times 1$ & \textbf{0.9050} &0.8239 & \textbf{0.4880} &\textbf{2.8178} &\textbf{0.3844}\\
    \midrule
    TinyHD-M$\times \frac{1}{2}$ & 0.9016 & 0.8272 & 0.4687 &2.6718 &0.3612\\
    \hspace{0.2cm} +TA & 0.9021 &0.8307 &0.4718 &2.6726 &0.3630\\
    \midrule
    TinyHD-M$\times \frac{1}{4}$ & 0.8980 &0.8294 &0.4487 &2.5257 &0.3438\\
    \hspace{0.2cm} +TA & 0.8999 & \textbf{0.8333} &0.4564 &2.5581 &0.3478\\
    \bottomrule
  \end{tabular}
  \end{subtable}
\end{table}

\section{Conclusions}
In this work, starting from the observation that different encoder-decoder architectures recognize specific video saliency patterns, we propose a heterogeneous multi-decoder architecture that leverages simpler versions of state-of-the-art decoding strategies to achieve high prediction accuracy at a fraction of the computational cost. We train our model in a multi-target knowledge distillation setting, where a hierarchical decoder is used as a teacher to supervise a matching internal decoder in our model and the output prediction; additionally, we employ semi-supervised learning on an unlabeled auxiliary dataset to further improve model generalization. Our model sets new state-of-the-art performance when employed in a multi-input/multi-output setting, while being significantly more efficient in terms of floating-point operations and number of parameters. We further push the limits of our model by applying a channel reduction procedure through multiple distillation steps and using the full-capacity student as a teacher, according to the ``teacher assistant'' paradigm. In the resulting model, the number of floating-point operations is approximately halved compared to the full-capacity version, and the number of parameters becomes as small as about 500k, taking about 2.4 MB storage space without compression.

\section*{Acknowledgments}
This publication has been financially supported by:
%has emanated from research conducted with the financial support of
Science Foundation Ireland (SFI) under grant number SFI/12/RC/2289\_P2;
Regione Sicilia, Italy, \emph{RehaStart} project (grant identifier: PO FESR 2014/2020, Azione 1.1.5, N. 08ME6201000222, CUP G79J18000610007); University of Catania, \emph{Piano della Ricerca di Ateneo}, 2020/2022, Linea 2D;
MIUR, Italy, Azione 1.2 ``Mobilità dei Ricercatori'' (grant identifier: Asse I, PON R\&I 2014-2020, id. AIM 1889410, CUP: E64I18002520007).

{\small
\bibliographystyle{ieee_fullname}
\bibliography{egbib}
}

\end{document}